\documentclass[runningheads]{llncs}
\usepackage[T1]{fontenc}
\usepackage[final]{microtype}
\usepackage[all]{nowidow}
\usepackage{graphicx}
\usepackage{wrapfig,booktabs,amsmath}

\usepackage{threeparttable}
\usepackage{multirow}
\usepackage{enumitem}
\usepackage{algorithm}
\usepackage{algpseudocode}
\newcommand{\bst}[1]{{\textbf{\textcolor{red}{#1}}}}
\newcommand{\subbst}[1]{\textcolor{blue}{\underline{{#1}}}}
\usepackage[table]{xcolor}
\newcommand{\scalea}[1]{\scalebox{0.78}{#1}}
\newcommand{\scaleb}[1]{\scalebox{0.8}{#1}}

\usepackage{amssymb}
\usepackage{bbding}
\usepackage{subfigure}

\usepackage{pifont}
\microtypesetup{protrusion=true,expansion=true}

\AtBeginDocument{\setlength{\parfillskip}{0pt plus .5\textwidth}}
\def \method{CoRe}
\begin{document}
\title{CoRe: Coherence and Relational Alignment for Multivariate Time Series Forecasting}
\titlerunning{CoRe for Multivariate Time Series Forecasting}

\author{
Xiaoyu Lin\inst{1}
\and
Huiran Duan\inst{2}
\and
Yining Liu\inst{3}
\and
Zhixiang Wu\inst{4}
\and
Chu Lin\inst{2}\\
\and
Lin Lu\inst{1,*}
}

\authorrunning{X. Lin et al.}

\institute{
College of Computer and Information Technology, China Three Gorges University, China\\
\and
City University of New York, USA\\
\and
University of California, Berkeley, USA\\
\and
Emory University, USA\\
\textsuperscript{*}Corresponding author. \email{lulin@ctgu.edu.cn}
}

\maketitle
\begin{abstract}
\looseness=-1 Direct forecasting has become a standard paradigm for multivariate time-series forecasting because it predicts the full future horizon in a single pass. However, its training objective is often still decomposed into pointwise errors such as MSE. Such objectives provide stable supervision, but they do not explicitly preserve the structure of the future trajectory: temporal coherence within each variable and relational consistency across variables can both be weakened. We propose \method, a model-agnostic learning objective for direct multivariate forecasting. \method\ replaces pointwise supervision with two output-space constraints: a frequency coherence loss that aligns predicted and target spectra, and a low-rank relational graph loss that matches sampled pairwise differences in a target-derived PCA subspace. The resulting objective introduces no trainable parameters and can be applied to existing forecasting backbones by changing only the loss. Experiments on standard benchmarks show that \method\ improves strong baselines, compares favorably with recent forecasting objectives, and remains effective across different backbones, datasets, and hyperparameter settings overall consistently.

\end{abstract}


\section{Introduction}

\looseness=-1 Deep learning is transforming a wide range of research domains\cite{zhao2026mis,li2026towards,wu2026roboalign,li2026rethinking,li2026comprehensive,lin2026cec,yang2026survey,xiao2026prototype,li2026mrmad}. Particularly rapid progress has been achieved in multimodal learning and high-performance AI\cite{Li2025Efficient,li2025frequency,li2026diff,feng2026mpq,feng2026s,kong2025token}. These advances create new possibilities for improving and extending scientific computing methods\cite{li2025pruning,li2025ddtime,xie2026symmetry,xiao2026points,xiao2026reversible}. Multivariate time-series forecasting (MTSF) is a core problem in energy systems, traffic networks, meteorology, and industrial monitoring. Given historical observations from multiple correlated variables, the task is to predict their future evolution over a target horizon. Recent forecasting models have improved accuracy with stronger temporal encoders, channel interaction modules, and scalable architectures~\cite{PatchTST,itransformer,Timesnet,11460474}. Among forecasting paradigms, direct forecasting (DF), which predicts the full horizon in one pass, is especially attractive because it avoids the error accumulation of iterative forecasting and supports efficient training and inference in practice.

Despite this architectural progress, the objective used to train DF models remains comparatively simple. Standard losses such as MSE decompose the future output into pointwise errors. This is convenient, but it does not explicitly match the structure of a multivariate future. Along the temporal dimension, future labels are autocorrelated and often contain periodic or long-range patterns. Along the variable dimension, channels are coupled by shared latent factors, such as neighboring sensors in traffic networks or correlated loads in electricity systems. A model can therefore reduce pointwise error while still producing predictions with distorted temporal profiles or weakened cross-variable geometry.

Recent work has begun to close this gap by revisiting the learning objective. Frequency-domain objectives such as FreDF~\cite{wang2025fredf} and dependency-aware objectives such as Time-o1~\cite{wang2025nipstimeo1} show that modeling future-label structure can improve direct forecasting without changing the backbone. However, these objectives mainly focus on temporal dependencies within each variable. For MTSF, the predicted future is also a joint multivariate object, and its cross-variable structure is typically left to the architecture rather than explicitly regularized at the output level.

We propose \textbf{\method} (\textbf{Co}herence and \textbf{Re}lational Alignment), a model-agnostic objective for direct multivariate forecasting. Instead of supervising each future value independently, \method\ trains the model with two complementary output-space constraints. A frequency coherence loss aligns predicted and target spectra along the future horizon, encouraging global temporal agreement. A low-rank relational graph loss projects predictions and targets into a target-derived PCA subspace and matches sampled pairwise differences between latent components, providing scalable supervision for cross-variable geometry. \method\ is a plug-in objective: it introduces no trainable parameters and can be used with representative Transformer, MLP, and CNN backbones by replacing their native loss.

Our contributions are summarized as follows:
\begin{itemize}[leftmargin=*, label=\textbullet]
    \item We identify an output-level objective gap in direct MTSF: pointwise losses do not explicitly preserve future-horizon coherence or cross-variable relational consistency in direct multivariate forecasting outputs.
    \item We propose \method, a model-agnostic objective that combines frequency coherence loss with efficient low-rank relational graph alignment without adding trainable parameters or architecture-level changes to the backbone.
    \item We validate \method\ on standard multivariate forecasting benchmarks, showing improvements over strong baselines, competitive objective-level methods, and multiple forecasting backbones across datasets and horizons.
\end{itemize}

\section{Related Work}

\subsection{MTSF Architectures and Cross-Variable Dependency Modeling}

MTSF requires modeling both temporal dynamics and dependencies among variables~\cite{qiu2026survey,ding2025dualsg}. Early statistical methods~\cite{Vector1993,Arima} have been largely extended by deep architectures, including RNNs~\cite{salinas2020deepar}, CNNs~\cite{Timesnet}, GNNs~\cite{HOU2026Graph}, MLPs~\cite{DLinear,wang2023timemixer,wang2024timemixer++}, and Transformers~\cite{11460474,li2025ddtimedatasetdistillationspectral,2026timemosaic,xu2026chainawareencodingmicroservicetrace,li2026gracegroundedreasoningadapter}. A central question in these models is how to represent cross-variable dependencies. Channel Dependence (CD) methods explicitly model cross-channel interactions~\cite{crossformer,itransformer}, while recent Channel Partiality (CP) methods filter noisy or redundant dependencies, e.g., DUET~\cite{qiu2025duet} with frequency-domain clustering and TimeFilter~\cite{hu2025timefilter} with patch-specific routing graphs.

These approaches mainly improve how the backbone extracts dependencies from historical observations. \method\ is complementary: it does not modify the backbone, but regularizes the structure of the predicted future itself. This distinction is important because a model can learn useful historical dependencies while still producing multivariate forecasts whose output geometry deviates from the target future horizon during forecasting.

\subsection{Direct Forecasting Objectives}

For multi-step forecasting, Iterative Forecasting (IF) recursively predicts future values and naturally preserves label autoregression, but it suffers from error accumulation~\cite{LSTNet,taieb2015bias}. Direct Forecasting (DF) predicts the entire horizon in one pass and is widely used for its efficiency and empirical strength~\cite{generative,PatchTST,itransformer}. However, standard DF training typically decomposes the future horizon into pointwise targets, weakening explicit supervision over future-label dependencies.

Recent learning objectives address this limitation from different angles. Shape-based objectives align temporal trajectories~\cite{GDTW,soft-dtw}, distribution-aware objectives improve probabilistic or likelihood-based supervision~\cite{wang2026iclrqdf,wang2025nipstimeo1,kmbdf,wang2026iclrdistdf}, and local objectives align steps or patches~\cite{TDAlign,patchloss}. FreDF~\cite{wang2025fredf} moves direct forecasting supervision into the frequency domain, and Time-o1~\cite{wang2025nipstimeo1} further studies objective-level modeling of future temporal dependencies. Related spectral models, such as Autoformer~\cite{Autoformer}, FedFormer~\cite{fedformer}, and FreTS~\cite{FreTS}, also show that frequency representations capture periodicity and long-range structure. Unlike these objectives, \method\ jointly regularizes future-horizon coherence and cross-variable relational geometry in the output space.

\section{Proposed Methodology}

\subsection{Problem Formulation and Objective Gap}

Let $\mathbf{X}\in\mathbb{R}^{B\times L\times D}$ denote a batch of historical observations, where $B$ is the batch size, $L$ is the look-back length, and $D$ is the number of variables. Given a forecasting backbone $f_{\theta}$, direct forecasting predicts the full future sequence $\hat{\mathbf{Y}}=f_{\theta}(\mathbf{X})\in\mathbb{R}^{B\times T\times D}$ for a target $\mathbf{Y}\in\mathbb{R}^{B\times T\times D}$.

Most DF models are trained with pointwise losses, e.g., MSE or MAE. These objectives are effective scalar criteria, but they do not explicitly constrain the joint structure of $\hat{\mathbf{Y}}$. We focus on two missing output-level constraints. First, the future horizon has temporal structure: neighboring and periodic labels are correlated, and matching isolated time steps may underconstrain the global trajectory. Second, the multivariate output has relational structure: variables should preserve relative patterns induced by shared latent factors. \method\ addresses this objective gap by replacing pointwise supervision with frequency coherence and low-rank relational graph alignment in the output space.

\begin{figure}[!t]
\centering
\includegraphics[width=0.92\linewidth]{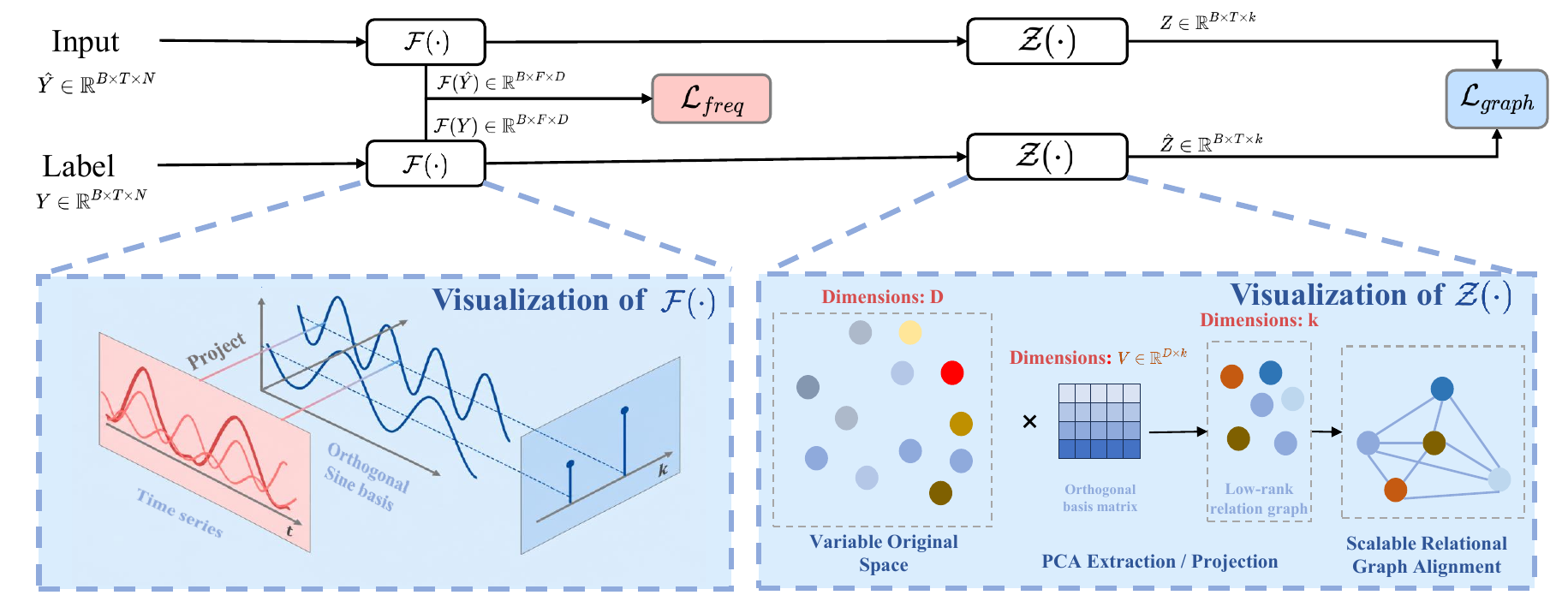}
\vspace{-0.8em}
\caption{Overall framework of \method, including frequency coherence and low-rank relational graph alignment.}
\label{fig:method_framework}
\vspace{-1.0em}
\end{figure}

Fig.~\ref{fig:method_framework} summarizes the overall framework of \method.

\subsection{Frequency Coherence Loss}

The first component aligns predictions and targets at the sequence level by applying the 1D real fast Fourier transform (RFFT) along the temporal dimension to both targets and predicted trajectories:
\begin{equation}
    \tilde{\mathbf{Y}}=\mathcal{F}(\mathbf{Y}), \quad
    \hat{\tilde{\mathbf{Y}}}=\mathcal{F}(\hat{\mathbf{Y}}),
\end{equation}
where $\tilde{\mathbf{Y}},\hat{\tilde{\mathbf{Y}}}\in\mathbb{C}^{B\times F\times D}$ and $F=\lfloor T/2\rfloor+1$.

The frequency coherence loss averages the complex-modulus distance over all batches, frequencies, and observed variables:
\begin{equation}
    \mathcal{L}_{freq}
    =
    \frac{1}{BFD}
    \sum_{b=1}^{B}\sum_{f=1}^{F}\sum_{d=1}^{D}
    \left|
    \hat{\tilde{\mathbf{Y}}}_{b,f,d}
    -
    \tilde{\mathbf{Y}}_{b,f,d}
    \right|.
\end{equation}
Because the loss compares spectra rather than individual time steps, it constrains both amplitude and phase information of the predicted trajectory. This encourages global temporal agreement, including periodicity and long-range evolution, while remaining independent of the forecasting backbone.

\subsection{Low-Rank Relational Graph Alignment}

The second component preserves cross-variable structure in the output space. Directly matching pairwise relations among all $D$ variables costs $\mathcal{O}(BTD^2)$ and can be sensitive to noisy high-dimensional correlations. Following recent efforts to explicitly preserve cross-variable relational structure \cite{ding2026cvloss}, we instead perform the alignment in a target-derived low-rank relational subspace.

For each training batch, we center the target sequence over the batch and temporal dimensions. Let $\bar{\mathbf{Y}}\in\mathbb{R}^{1\times 1\times D}$ be the target mean, and let $\mathbf{V}\in\mathbb{R}^{D\times k}$ contain the top-$k$ orthonormal principal directions computed from the target-side covariance over variables. We project predictions and targets with the same basis:
\begin{equation}
    \mathbf{Z} = (\mathbf{Y}-\bar{\mathbf{Y}})\mathbf{V}, \quad
    \hat{\mathbf{Z}} = (\hat{\mathbf{Y}}-\bar{\mathbf{Y}})\mathbf{V},
\end{equation}
where $\mathbf{Z},\hat{\mathbf{Z}}\in\mathbb{R}^{B\times T\times k}$. Sharing the target-derived centering and projection gives both outputs a common coordinate system for relational comparison.

We then sample unordered latent component pairs $\mathcal{E}\subset\{1,\ldots,k\}^2$ and match first-order differences between each sampled pair:
\begin{equation}
    \mathcal{L}_{graph}
    =
    \frac{1}{BT|\mathcal{E}|}
    \sum_{b=1}^{B}\sum_{t=1}^{T}
    \sum_{(i,j)\in\mathcal{E}}
    \left|
    \left(\hat{\mathbf{Z}}_{b,t,i}-\hat{\mathbf{Z}}_{b,t,j}\right)
    -
    \left(\mathbf{Z}_{b,t,i}-\mathbf{Z}_{b,t,j}\right)
    \right|.
\end{equation}
This LR-Diff loss penalizes deviations in relative latent responses rather than isolated channel values. After projection, random edge sampling reduces the pairwise comparison cost from $\mathcal{O}(BTk^2)$ to $\mathcal{O}(BT|\mathcal{E}|)$, making the relational constraint scalable for datasets with many variables.

\subsection{Overall Objective}

\method\ trains the forecasting backbone with a weighted objective that combines the two structural terms used by CoRe:
\begin{equation}
    \mathcal{L}_{total}
    =
    \alpha \mathcal{L}_{graph}
    +
    (1-\alpha)\mathcal{L}_{freq},
\end{equation}
where $\alpha\in[0,1]$ controls the trade-off between relational alignment and frequency coherence throughout the objective optimization process.

During training, the backbone produces $\hat{\mathbf{Y}}$ from the historical input, and \method\ computes both losses directly on the predicted and target future sequences. MSE and MAE are not used as training losses in \method; they are kept only as evaluation metrics for comparability. Since \method\ changes only the objective, it can be applied to existing forecasting architectures without adding trainable parameters or modifying the model design. Its extra cost stays in the output space during training: the RFFT over the horizon costs $\mathcal{O}(BDT\log T)$, and edge sampling reduces the latent pairwise comparison from $\mathcal{O}(BTk^2)$ to $\mathcal{O}(BT|\mathcal{E}|)$, while inference remains unchanged.

\section{Experiments}


\subsection{Setup}

\paragraph{Datasets.}
We use standard multivariate forecasting benchmarks: ETT~\cite{Informer}, Weather, ECL, Traffic~\cite{Autoformer}, and PEMS~\cite{SCINet}. These datasets cover electricity transformers, meteorological measurements, electricity consumption, freeway traffic occupancy, and traffic sensor networks, and therefore include both low-dimensional and high-dimensional forecasting settings. Following common protocols~\cite{itransformer}, all datasets are split chronologically into training, validation, and test sets to avoid temporal leakage. The input length is fixed to 96 for ETT, Weather, ECL, and Traffic, with prediction lengths $\{96,192,336,720\}$; for PEMS, the prediction lengths are $\{12,24,36,48\}$.

\paragraph{Baselines.} We compare with representative forecasting architectures from three families: Transformer-based models (PatchTST~\cite{PatchTST}, FEDformer~\cite{fedformer}, iTransformer~\cite{itransformer}), MLP-based models (DLinear~\cite{DLinear}, TiDE~\cite{das2023long}, FreTS~\cite{FreTS}), and other competitive architectures (TimesNet~\cite{Timesnet}, MICN~\cite{MICN}). Since CoRe is an objective rather than a new backbone, we also compare with objective-level baselines, including FreDF~\cite{wang2025fredf} and Time-o1~\cite{wang2025nipstimeo1}, in the dedicated learning-objective study.

\paragraph{Implementation.} Baselines are reproduced using the scripts from~\cite{itransformer} and trained with Adam~\cite{Adam} under the standard MSE objective. We follow the same chronological splits and evaluation protocol for all methods, and disable the drop-last trick during testing following~\cite{qiutfb}. When applying CoRe to an existing backbone, we keep the benchmark hyperparameters whenever possible, so that the comparison mainly reflects the effect of the training objective. We tune only the learning rate and two CoRe-specific parameters: the loss weight $\alpha$ and PCA dimension $k$. The PCA basis is estimated from the target side of each mini-batch, and the same target-derived basis is used to project both predictions and targets. Latent component pairs are sampled uniformly without replacement for the relational graph loss, and the selected hyperparameters are chosen on the validation split. Experiments are conducted on Intel(R) Xeon(R) Gold 6248R CPUs and 8 NVIDIA RTX 3090 GPUs.

\begin{table}
  \caption{Long-term forecasting results with input length 96 following~\cite{itransformer}. \bst{Bold} denotes the best result and \subbst{underlined} denotes the second-best result for each metric.}\label{tab:longterm_app}
  \vspace{-5pt}
  \renewcommand{\arraystretch}{0.95}
  \setlength{\tabcolsep}{6pt} \scriptsize
  \centering
  \renewcommand{\multirowsetup}{\centering}
  \begin{threeparttable}
  \resizebox{0.9\columnwidth}{!}{%
  \begin{tabular}{c|c|cc|cc|cc|cc|cc|cc|cc|cc|cc|cc|cc}
    \toprule
    \multicolumn{2}{l}{\multirow{2}{*}{\rotatebox{0}{\scaleb{Models}}}} & 
    \multicolumn{2}{c}{\rotatebox{0}{\scaleb{\textbf{CoRe}}}} &
    \multicolumn{2}{c}{\rotatebox{0}{\scaleb{{FreDF}}}} &
    \multicolumn{2}{c}{\rotatebox{0}{\scaleb{iTransformer}}} &
    \multicolumn{2}{c}{\rotatebox{0}{\scaleb{FreTS}}} &
    \multicolumn{2}{c}{\rotatebox{0}{\scaleb{TimesNet}}} &
    \multicolumn{2}{c}{\rotatebox{0}{\scaleb{MICN}}} &
    \multicolumn{2}{c}{\rotatebox{0}{\scaleb{TiDE}}} &
    \multicolumn{2}{c}{\rotatebox{0}{\scaleb{DLinear}}} &
    \multicolumn{2}{c}{\rotatebox{0}{\scaleb{PatchTST}}} &
    \multicolumn{2}{c}{\rotatebox{0}{\scaleb{FEDformer}}} &
    \multicolumn{2}{c}{\rotatebox{0}{\scaleb{Transformer}}} \\
    \multicolumn{2}{c}{} &
    \multicolumn{2}{c}{\scaleb{\textbf{(Ours)}}} &
    \multicolumn{2}{c}{\scaleb{\textbf{(2025)}}} & 
    \multicolumn{2}{c}{\scaleb{(2024)}} & 
    \multicolumn{2}{c}{\scaleb{(2023)}} & 
    \multicolumn{2}{c}{\scaleb{(2023)}} &
    \multicolumn{2}{c}{\scaleb{(2023)}} & 
    \multicolumn{2}{c}{\scaleb{(2023)}} & 
    \multicolumn{2}{c}{\scaleb{(2023)}} & 
    \multicolumn{2}{c}{\scaleb{(2023)}} &
    \multicolumn{2}{c}{\scaleb{(2022)}} &
    \multicolumn{2}{c}{\scaleb{(2017)}} \\
    \cmidrule(lr){3-4} \cmidrule(lr){5-6}\cmidrule(lr){7-8} \cmidrule(lr){9-10}\cmidrule(lr){11-12} \cmidrule(lr){13-14} \cmidrule(lr){15-16} \cmidrule(lr){17-18} \cmidrule(lr){19-20} \cmidrule(lr){21-22} \cmidrule(lr){23-24}
    \multicolumn{2}{l}{\rotatebox{0}{\scaleb{Metrics}}}  & \scalea{MSE} & \scalea{MAE}  & \scalea{MSE} & \scalea{MAE}  & \scalea{MSE} & \scalea{MAE}  & \scalea{MSE} & \scalea{MAE}  & \scalea{MSE} & \scalea{MAE}  & \scalea{MSE} & \scalea{MAE} & \scalea{MSE} & \scalea{MAE} & \scalea{MSE} & \scalea{MAE} & \scalea{MSE} & \scalea{MAE} & \scalea{MSE} & \scalea{MAE} & \scalea{MSE} & \scalea{MAE} \\
    \toprule

    \multirow{5}{*}{{\rotatebox{90}{\scalebox{0.95}{ETTm1}}}}
    & \scalea{96} & \scalea{\bst{0.308}} & \scalea{\bst{0.346}} & \scalea{\subbst{0.324}} & \scalea{\subbst{0.362}} & \scalea{0.346} & \scalea{0.379} & \scalea{0.339} & \scalea{0.374} & \scalea{0.338} & \scalea{0.379} & \scalea{0.318} & \scalea{0.366} & \scalea{0.364} & \scalea{0.387} & \scalea{0.345} & \scalea{0.372} & \scalea{0.325} & \scalea{0.364} & \scalea{0.389} & \scalea{0.427} & \scalea{0.591} & \scalea{0.549} \\
    & \scalea{192} & \scalea{\bst{0.360}} & \scalea{\bst{0.373}} & \scalea{0.373} & \scalea{0.385} & \scalea{0.392} & \scalea{0.400} & \scalea{0.382} & \scalea{0.397} & \scalea{0.389} & \scalea{0.400} & \scalea{0.364} & \scalea{0.396} & \scalea{0.398} & \scalea{0.404} & \scalea{0.381} & \scalea{0.390} & \scalea{\subbst{0.363}} & \scalea{\subbst{0.383}} & \scalea{0.402} & \scalea{0.431} & \scalea{0.704} & \scalea{0.629} \\
    & \scalea{336} & \scalea{\bst{0.393}} & \scalea{\bst{0.395}} & \scalea{0.402} & \scalea{\subbst{0.404}} & \scalea{0.427} & \scalea{0.422} & \scalea{0.421} & \scalea{0.426} & \scalea{0.429} & \scalea{0.428} & \scalea{\subbst{0.398}} & \scalea{0.428} & \scalea{0.428} & \scalea{0.425} & \scalea{0.414} & \scalea{0.414} & \scalea{0.404} & \scalea{0.413} & \scalea{0.438} & \scalea{0.451} & \scalea{1.171} & \scalea{0.861} \\
    & \scalea{720} & \scalea{\bst{0.462}} & \scalea{\bst{0.434}} & \scalea{0.469} & \scalea{0.444} & \scalea{0.494} & \scalea{0.461} & \scalea{0.485} & \scalea{0.462} & \scalea{0.495} & \scalea{0.464} & \scalea{0.514} & \scalea{0.501} & \scalea{0.487} & \scalea{0.461} & \scalea{0.473} & \scalea{0.451} & \scalea{\subbst{0.463}} & \scalea{\subbst{0.442}} & \scalea{0.529} & \scalea{0.498} & \scalea{1.307} & \scalea{0.893} \\
    \cmidrule(lr){2-24}
    & \scalea{Avg} & \scalea{\bst{0.381}} & \scalea{\bst{0.387}} & \scalea{0.392} & \scalea{\subbst{0.399}} & \scalea{0.415} & \scalea{0.416} & \scalea{0.407} & \scalea{0.415} & \scalea{0.413} & \scalea{0.418} & \scalea{\subbst{0.399}} & \scalea{0.423} & \scalea{0.419} & \scalea{0.419} & \scalea{0.404} & \scalea{\subbst{0.407}} & \scalea{0.389} & \scalea{0.400} & \scalea{0.440} & \scalea{0.451} & \scalea{0.943} & \scalea{0.733} \\
    \midrule

    \multirow{5}{*}{{\rotatebox{90}{\scalebox{0.95}{ETTm2}}}}
    & \scalea{96}  & \scalea{\bst{0.172}} & \scalea{\bst{0.250}} & \scalea{\subbst{0.173}} & \scalea{\subbst{0.252}} & \scalea{0.184} & \scalea{0.266} & \scalea{0.190} & \scalea{0.282} & \scalea{0.185} & \scalea{0.264} & \scalea{0.178} & \scalea{0.275} & \scalea{0.207} & \scalea{0.305} & \scalea{0.195} & \scalea{0.294} & \scalea{0.180} & \scalea{0.266} & \scalea{0.194} & \scalea{0.284} & \scalea{0.317} & \scalea{0.408} \\
    & \scalea{192} & \scalea{\bst{0.237}} & \scalea{\bst{0.294}} & \scalea{0.241} & \scalea{\subbst{0.298}} & \scalea{0.257} & \scalea{0.315} & \scalea{0.260} & \scalea{0.329} & \scalea{0.254} & \scalea{0.307} & \scalea{\subbst{0.240}} & \scalea{0.317} & \scalea{0.290} & \scalea{0.364} & \scalea{0.283} & \scalea{0.359} & \scalea{0.285} & \scalea{0.339} & \scalea{0.264} & \scalea{0.324} & \scalea{1.069} & \scalea{0.758} \\
    & \scalea{336} & \scalea{\bst{0.296}} & \scalea{\bst{0.333}} & \scalea{\subbst{0.298}} & \scalea{\subbst{0.334}} & \scalea{0.315} & \scalea{0.351} & \scalea{0.373} & \scalea{0.405} & \scalea{0.314} & \scalea{0.345} & \scalea{0.299} & \scalea{0.354} & \scalea{0.377} & \scalea{0.422} & \scalea{0.384} & \scalea{0.427} & \scalea{0.309} & \scalea{0.347} & \scalea{0.319} & \scalea{0.359} & \scalea{1.325} & \scalea{0.869} \\
    & \scalea{720} & \scalea{\bst{0.394}} & \scalea{\bst{0.391}} & \scalea{\subbst{0.398}} & \scalea{\subbst{0.393}} & \scalea{0.419} & \scalea{0.409} & \scalea{0.517} & \scalea{0.499} & \scalea{0.434} & \scalea{0.413} & \scalea{0.482} & \scalea{0.479} & \scalea{0.558} & \scalea{0.524} & \scalea{0.516} & \scalea{0.502} & \scalea{0.437} & \scalea{0.422} & \scalea{0.430} & \scalea{0.424} & \scalea{2.576} & \scalea{1.223} \\
    \cmidrule(lr){2-24}
    & \scalea{Avg} & \scalea{\bst{0.275}} & \scalea{\bst{0.317}} & \scalea{\subbst{0.278}} & \scalea{\subbst{0.319}} & \scalea{0.294} & \scalea{0.335} & \scalea{0.335} & \scalea{0.379} & \scalea{0.297} & \scalea{0.332} & \scalea{0.300} & \scalea{0.356} & \scalea{0.358} & \scalea{0.404} & \scalea{0.344} & \scalea{0.396} & \scalea{0.303} & \scalea{0.344} & \scalea{0.302} & \scalea{0.348} & \scalea{1.322} & \scalea{0.814} \\
    \midrule

    \multirow{5}{*}{\rotatebox{90}{{\scalebox{0.95}{ETTh1}}}}
    & \scalea{96} & \scalea{\bst{0.376}} & \scalea{\bst{0.391}} & \scalea{0.382} & \scalea{0.400} & \scalea{0.390} & \scalea{0.410} & \scalea{0.399} & \scalea{0.412} & \scalea{0.422} & \scalea{0.433} & \scalea{0.383} & \scalea{0.418} & \scalea{0.479} & \scalea{0.464} & \scalea{0.396} & \scalea{0.410} & \scalea{0.381} & \scalea{\subbst{0.400}} & \scalea{\subbst{0.377}} & \scalea{0.418} & \scalea{0.796} & \scalea{0.691} \\
    & \scalea{192} & \scalea{\subbst{0.424}} & \scalea{\bst{0.425}} & \scalea{0.430} & \scalea{\subbst{0.427}} & \scalea{0.443} & \scalea{0.441} & \scalea{0.453} & \scalea{0.443} & \scalea{0.465} & \scalea{0.457} & \scalea{0.500} & \scalea{0.491} & \scalea{0.521} & \scalea{0.503} & \scalea{0.449} & \scalea{0.444} & \scalea{0.450} & \scalea{0.443} & \scalea{\bst{0.421}} & \scalea{0.445} & \scalea{0.813} & \scalea{0.699} \\
    & \scalea{336} & \scalea{\bst{0.463}} & \scalea{\bst{0.446}} & \scalea{0.474} & \scalea{\subbst{0.451}} & \scalea{0.480} & \scalea{0.457} & \scalea{0.503} & \scalea{0.475} & \scalea{0.492} & \scalea{0.470} & \scalea{0.546} & \scalea{0.530} & \scalea{0.659} & \scalea{0.603} & \scalea{0.487} & \scalea{0.465} & \scalea{0.501} & \scalea{0.470} & \scalea{\subbst{0.468}} & \scalea{0.472} & \scalea{1.181} & \scalea{0.876} \\
    & \scalea{720} & \scalea{0.473} & \scalea{\subbst{0.471}} & \scalea{\bst{0.463}} & \scalea{\bst{0.462}} & \scalea{0.484} & \scalea{0.479} & \scalea{0.596} & \scalea{0.565} & \scalea{0.532} & \scalea{0.502} & \scalea{0.671} & \scalea{0.620} & \scalea{0.893} & \scalea{0.736} & \scalea{0.516} & \scalea{0.513} & \scalea{\bst{0.463}} & \scalea{0.492} & \scalea{0.500} & \scalea{0.493} & \scalea{1.182} & \scalea{0.885} \\
    \cmidrule(lr){2-24}
    & \scalea{Avg} & \scalea{\bst{0.434}} & \scalea{\bst{0.433}} & \scalea{0.437} & \scalea{\subbst{0.435}} & \scalea{0.449} & \scalea{0.447} & \scalea{0.488} & \scalea{0.474} & \scalea{0.478} & \scalea{0.466} & \scalea{0.525} & \scalea{0.515} & \scalea{0.628} & \scalea{0.574} & \scalea{0.462} & \scalea{0.458} & \scalea{0.459} & \scalea{0.451} & \scalea{\subbst{0.441}} & \scalea{0.457} & \scalea{0.993} & \scalea{0.788} \\
    \midrule

    \multirow{5}{*}{\rotatebox{90}{{\scalebox{0.95}{ETTh2}}}}
    & \scalea{96}  & \scalea{\bst{0.285}} & \scalea{\bst{0.334}} & \scalea{\subbst{0.289}} & \scalea{\subbst{0.337}} & \scalea{0.301} & \scalea{0.349} & \scalea{0.350} & \scalea{0.403} & \scalea{0.320} & \scalea{0.364} & \scalea{0.361} & \scalea{0.404} & \scalea{0.400} & \scalea{0.440} & \scalea{0.343} & \scalea{0.396} & \scalea{0.299} & \scalea{0.349} & \scalea{0.347} & \scalea{0.391} & \scalea{2.072} & \scalea{1.140} \\
    & \scalea{192} & \scalea{\bst{0.363}} & \scalea{\bst{0.383}} & \scalea{\bst{0.363}} & \scalea{\subbst{0.385}} & \scalea{0.382} & \scalea{0.402} & \scalea{0.472} & \scalea{0.475} & \scalea{0.409} & \scalea{0.417} & \scalea{0.495} & \scalea{0.490} & \scalea{0.528} & \scalea{0.509} & \scalea{0.473} & \scalea{0.474} & \scalea{0.383} & \scalea{0.404} & \scalea{0.430} & \scalea{0.443} & \scalea{5.081} & \scalea{1.814} \\
    & \scalea{336} & \scalea{\bst{0.409}} & \scalea{\bst{0.420}} & \scalea{\subbst{0.419}} & \scalea{\subbst{0.426}} & \scalea{0.430} & \scalea{0.434} & \scalea{0.564} & \scalea{0.528} & \scalea{0.449} & \scalea{0.451} & \scalea{0.671} & \scalea{0.588} & \scalea{0.643} & \scalea{0.571} & \scalea{0.603} & \scalea{0.546} & \scalea{0.439} & \scalea{0.444} & \scalea{0.469} & \scalea{0.475} & \scalea{3.564} & \scalea{1.475} \\
    & \scalea{720} & \scalea{\bst{0.413}} & \scalea{\bst{0.435}} & \scalea{\subbst{0.415}} & \scalea{\subbst{0.437}} & \scalea{0.447} & \scalea{0.455} & \scalea{0.815} & \scalea{0.654} & \scalea{0.473} & \scalea{0.474} & \scalea{0.968} & \scalea{0.712} & \scalea{0.874} & \scalea{0.679} & \scalea{0.812} & \scalea{0.650} & \scalea{0.438} & \scalea{0.455} & \scalea{0.473} & \scalea{0.480} & \scalea{2.469} & \scalea{1.247} \\
    \cmidrule(lr){2-24}
    & \scalea{Avg} & \scalea{\bst{0.367}} & \scalea{\bst{0.393}} & \scalea{\subbst{0.371}} & \scalea{\subbst{0.396}} & \scalea{0.390} & \scalea{0.410} & \scalea{0.550} & \scalea{0.515} & \scalea{0.413} & \scalea{0.426} & \scalea{0.624} & \scalea{0.549} & \scalea{0.611} & \scalea{0.550} & \scalea{0.558} & \scalea{0.516} & \scalea{0.390} & \scalea{0.413} & \scalea{0.430} & \scalea{0.447} & \scalea{3.296} & \scalea{1.419} \\
    \midrule

    \multirow{5}{*}{{\rotatebox{90}{\scalebox{0.95}{ECL}}}} 
    & \scalea{96} & \scalea{\bst{0.143}} & \scalea{\bst{0.231}} & \scalea{\subbst{0.144}} & \scalea{\subbst{0.233}} & \scalea{0.148} & \scalea{0.239} & \scalea{0.189} & \scalea{0.277} & \scalea{0.171} & \scalea{0.273} & \scalea{0.168} & \scalea{0.280} & \scalea{0.237} & \scalea{0.329} & \scalea{0.210} & \scalea{0.302} & \scalea{0.170} & \scalea{0.264} & \scalea{0.200} & \scalea{0.315} & \scalea{0.252} & \scalea{0.352} \\
    & \scalea{192} & \scalea{\bst{0.158}} & \scalea{\bst{0.246}} & \scalea{\subbst{0.159}} & \scalea{\subbst{0.247}} & \scalea{0.167} & \scalea{0.258} & \scalea{0.193} & \scalea{0.282} & \scalea{0.188} & \scalea{0.289} & \scalea{0.177} & \scalea{0.289} & \scalea{0.236} & \scalea{0.330} & \scalea{0.210} & \scalea{0.305} & \scalea{0.179} & \scalea{0.273} & \scalea{0.207} & \scalea{0.322} & \scalea{0.266} & \scalea{0.364} \\
    & \scalea{336} & \scalea{\bst{0.171}} & \scalea{\bst{0.261}} & \scalea{\subbst{0.172}} & \scalea{\subbst{0.263}} & \scalea{0.179} & \scalea{0.272} & \scalea{0.207} & \scalea{0.296} & \scalea{0.208} & \scalea{0.304} & \scalea{0.185} & \scalea{0.296} & \scalea{0.249} & \scalea{0.344} & \scalea{0.223} & \scalea{0.319} & \scalea{0.195} & \scalea{0.288} & \scalea{0.226} & \scalea{0.340} & \scalea{0.292} & \scalea{0.383} \\
    & \scalea{720} & \scalea{\bst{0.202}} & \scalea{\bst{0.289}} & \scalea{\subbst{0.204}} & \scalea{\subbst{0.294}} & \scalea{0.209} & \scalea{0.298} & \scalea{0.245} & \scalea{0.332} & \scalea{0.289} & \scalea{0.363} & \scalea{0.218} & \scalea{0.323} & \scalea{0.284} & \scalea{0.373} & \scalea{0.258} & \scalea{0.350} & \scalea{0.234} & \scalea{0.320} & \scalea{0.282} & \scalea{0.379} & \scalea{0.287} & \scalea{0.371} \\
    \cmidrule(lr){2-24}
    & \scalea{Avg} & \scalea{\bst{0.168}} & \scalea{\bst{0.257}} & \scalea{\subbst{0.170}} & \scalea{\subbst{0.259}} & \scalea{0.176} & \scalea{0.267} & \scalea{0.209} & \scalea{0.297} & \scalea{0.214} & \scalea{0.307} & \scalea{0.187} & \scalea{0.297} & \scalea{0.251} & \scalea{0.344} & \scalea{0.225} & \scalea{0.319} & \scalea{0.195} & \scalea{0.286} & \scalea{0.229} & \scalea{0.339} & \scalea{0.274} & \scalea{0.367} \\
    \midrule

    \multirow{5}{*}{{\rotatebox{90}{\scalebox{0.95}{Traffic}}}} 
    & \scalea{96} & \scalea{\bst{0.387}} & \scalea{\bst{0.248}} & \scalea{\subbst{0.391}} & \scalea{\subbst{0.265}} & \scalea{0.397} & \scalea{0.272} & \scalea{0.528} & \scalea{0.341} & \scalea{0.504} & \scalea{0.298} & \scalea{0.609} & \scalea{0.317} & \scalea{0.805} & \scalea{0.493} & \scalea{0.697} & \scalea{0.429} & \scalea{0.444} & \scalea{0.284} & \scalea{0.577} & \scalea{0.362} & \scalea{0.686} & \scalea{0.385} \\
    & \scalea{192} & \scalea{\bst{0.405}} & \scalea{\bst{0.269}} & \scalea{\subbst{0.410}} & \scalea{\subbst{0.273}} & \scalea{0.418} & \scalea{0.279} & \scalea{0.531} & \scalea{0.338} & \scalea{0.526} & \scalea{0.305} & \scalea{0.621} & \scalea{0.328} & \scalea{0.756} & \scalea{0.474} & \scalea{0.647} & \scalea{0.407} & \scalea{0.454} & \scalea{0.291} & \scalea{0.603} & \scalea{0.372} & \scalea{0.679} & \scalea{0.377} \\
    & \scalea{336} & \scalea{\bst{0.419}} & \scalea{\bst{0.277}} & \scalea{\subbst{0.424}} & \scalea{\subbst{0.280}} & \scalea{0.432} & \scalea{0.286} & \scalea{0.551} & \scalea{0.345} & \scalea{0.540} & \scalea{0.310} & \scalea{0.641} & \scalea{0.342} & \scalea{0.762} & \scalea{0.477} & \scalea{0.653} & \scalea{0.410} & \scalea{0.469} & \scalea{0.298} & \scalea{0.615} & \scalea{0.378} & \scalea{0.663} & \scalea{0.361} \\
    & \scalea{720} & \scalea{\bst{0.452}} & \scalea{\bst{0.292}} & \scalea{\subbst{0.460}} & \scalea{\subbst{0.298}} & \scalea{0.467} & \scalea{0.305} & \scalea{0.598} & \scalea{0.367} & \scalea{0.570} & \scalea{0.324} & \scalea{0.671} & \scalea{0.354} & \scalea{0.719} & \scalea{0.449} & \scalea{0.694} & \scalea{0.429} & \scalea{0.506} & \scalea{0.319} & \scalea{0.649} & \scalea{0.403} & \scalea{0.693} & \scalea{0.381} \\
    \cmidrule(lr){2-24}
    & \scalea{Avg} & \scalea{\bst{0.416}} & \scalea{\bst{0.272}} & \scalea{\subbst{0.421}} & \scalea{\subbst{0.279}} & \scalea{0.428} & \scalea{0.286} & \scalea{0.552} & \scalea{0.348} & \scalea{0.535} & \scalea{0.309} & \scalea{0.636} & \scalea{0.335} & \scalea{0.760} & \scalea{0.473} & \scalea{0.673} & \scalea{0.419} & \scalea{0.468} & \scalea{0.298} & \scalea{0.611} & \scalea{0.379} & \scalea{0.680} & \scalea{0.376} \\
    \midrule

    \multirow{5}{*}{{\rotatebox{90}{\scalebox{0.95}{Weather}}}}
    & \scalea{96}  & \scalea{\bst{0.163}} & \scalea{\bst{0.202}} & \scalea{\subbst{0.164}} & \scalea{\bst{0.202}} & \scalea{0.201} & \scalea{0.247} & \scalea{0.184} & \scalea{0.239} & \scalea{0.178} & \scalea{0.226} & \scalea{0.182} & \scalea{0.250} & \scalea{0.202} & \scalea{0.261} & \scalea{0.197} & \scalea{0.259} & \scalea{0.189} & \scalea{0.230} & \scalea{0.221} & \scalea{0.304} & \scalea{0.332} & \scalea{0.383} \\
    & \scalea{192} & \scalea{\bst{0.216}} & \scalea{\bst{0.250}} & \scalea{\subbst{0.220}} & \scalea{\subbst{0.253}} & \scalea{0.250} & \scalea{0.283} & \scalea{0.223} & \scalea{0.275} & \scalea{0.227} & \scalea{0.266} & \scalea{0.234} & \scalea{0.301} & \scalea{0.242} & \scalea{0.298} & \scalea{0.236} & \scalea{0.294} & \scalea{0.228} & \scalea{0.262} & \scalea{0.275} & \scalea{0.345} & \scalea{0.634} & \scalea{0.539} \\
    & \scalea{336} & \scalea{\subbst{0.275}} & \scalea{\bst{0.293}} & \scalea{\subbst{0.275}} & \scalea{\subbst{0.294}} & \scalea{0.302} & \scalea{0.317} & \scalea{0.272} & \scalea{0.316} & \scalea{0.283} & \scalea{0.305} & \scalea{\bst{0.268}} & \scalea{0.325} & \scalea{0.287} & \scalea{0.335} & \scalea{0.282} & \scalea{0.332} & \scalea{0.288} & \scalea{0.305} & \scalea{0.338} & \scalea{0.379} & \scalea{0.656} & \scalea{0.579} \\
    & \scalea{720} & \scalea{\subbst{0.353}} & \scalea{\bst{0.345}} & \scalea{0.356} & \scalea{\subbst{0.347}} & \scalea{0.370} & \scalea{0.362} & \scalea{\bst{0.340}} & \scalea{0.363} & \scalea{0.359} & \scalea{0.355} & \scalea{0.361} & \scalea{0.399} & \scalea{0.351} & \scalea{0.386} & \scalea{0.347} & \scalea{0.384} & \scalea{0.362} & \scalea{0.354} & \scalea{0.408} & \scalea{0.418} & \scalea{0.908} & \scalea{0.706} \\
    \cmidrule(lr){2-24}
    & \scalea{Avg}  & \scalea{\bst{0.251}} & \scalea{\bst{0.272}} & \scalea{\subbst{0.254}} & \scalea{\subbst{0.274}} & \scalea{0.281} & \scalea{0.302} & \scalea{0.255} & \scalea{0.299} & \scalea{0.262} & \scalea{0.288} & \scalea{0.261} & \scalea{0.319} & \scalea{0.271} & \scalea{0.320} & \scalea{0.265} & \scalea{0.317} & \scalea{0.267} & \scalea{0.288} & \scalea{0.311} & \scalea{0.361} & \scalea{0.632} & \scalea{0.552} \\
    \midrule

    \multirow{5}{*}{{\rotatebox{90}{\scalebox{0.95}{PEMS03}}}}
    & \scalea{12}  & \scalea{\bst{0.066}} & \scalea{\bst{0.169}} & \scalea{\subbst{0.068}} & \scalea{\subbst{0.172}} & \scalea{0.069} & \scalea{0.175} & \scalea{0.083} & \scalea{0.194} & \scalea{0.082} & \scalea{0.188} & \scalea{0.087} & \scalea{0.203} & \scalea{0.117} & \scalea{0.225} & \scalea{0.122} & \scalea{0.245} & \scalea{0.092} & \scalea{0.210} & \scalea{0.123} & \scalea{0.248} & \scalea{0.107} & \scalea{0.209} \\
    & \scalea{24}  & \scalea{\subbst{0.093}} & \scalea{\subbst{0.202}} & \scalea{0.096} & \scalea{0.205} & \scalea{0.098} & \scalea{0.210} & \scalea{0.127} & \scalea{0.241} & \scalea{0.110} & \scalea{0.216} & \scalea{\bst{0.086}} & \scalea{\bst{0.198}} & \scalea{0.233} & \scalea{0.320} & \scalea{0.202} & \scalea{0.320} & \scalea{0.144} & \scalea{0.263} & \scalea{0.160} & \scalea{0.287} & \scalea{0.121} & \scalea{0.227} \\
    & \scalea{36}  & \scalea{\subbst{0.122}} & \scalea{\subbst{0.232}} & \scalea{0.128} & \scalea{0.240} & \scalea{0.131} & \scalea{0.243} & \scalea{0.169} & \scalea{0.281} & \scalea{0.133} & \scalea{0.236} & \scalea{\bst{0.105}} & \scalea{\bst{0.220}} & \scalea{0.380} & \scalea{0.422} & \scalea{0.275} & \scalea{0.382} & \scalea{0.200} & \scalea{0.309} & \scalea{0.191} & \scalea{0.321} & \scalea{0.133} & \scalea{0.243} \\
    & \scalea{48}  & \scalea{0.155} & \scalea{0.264} & \scalea{0.161} & \scalea{0.269} & \scalea{0.164} & \scalea{0.275} & \scalea{0.204} & \scalea{0.311} & \scalea{\subbst{0.146}} & \scalea{\subbst{0.251}} & \scalea{\bst{0.120}} & \scalea{\bst{0.235}} & \scalea{0.536} & \scalea{0.511} & \scalea{0.335} & \scalea{0.429} & \scalea{0.245} & \scalea{0.344} & \scalea{0.223} & \scalea{0.350} & \scalea{0.144} & \scalea{0.253} \\
    \cmidrule(lr){2-24}
    & \scalea{Avg}  & \scalea{\subbst{0.109}} & \scalea{\subbst{0.217}} & \scalea{0.113} & \scalea{0.219} & \scalea{0.116} & \scalea{0.226} & \scalea{0.146} & \scalea{0.257} & \scalea{0.118} & \scalea{0.223} & \scalea{\bst{0.099}} & \scalea{\bst{0.214}} & \scalea{0.316} & \scalea{0.370} & \scalea{0.233} & \scalea{0.344} & \scalea{0.170} & \scalea{0.282} & \scalea{0.174} & \scalea{0.302} & \scalea{0.126} & \scalea{0.233} \\
    \midrule

    \multirow{5}{*}{{\rotatebox{90}{\scalebox{0.95}{PEMS08}}}}
    & \scalea{12}  & \scalea{\bst{0.078}} & \scalea{\bst{0.179}} & \scalea{\subbst{0.080}} & \scalea{\subbst{0.182}} & \scalea{0.085} & \scalea{0.189} & \scalea{0.095} & \scalea{0.204} & \scalea{0.110} & \scalea{0.209} & \scalea{2.193} & \scalea{0.871} & \scalea{0.121} & \scalea{0.231} & \scalea{0.152} & \scalea{0.274} & \scalea{0.106} & \scalea{0.223} & \scalea{0.175} & \scalea{0.275} & \scalea{0.213} & \scalea{0.236} \\
    & \scalea{24}  & \scalea{\bst{0.113}} & \scalea{\bst{0.213}} & \scalea{\subbst{0.118}} & \scalea{\subbst{0.220}} & \scalea{0.131} & \scalea{0.236} & \scalea{0.150} & \scalea{0.259} & \scalea{0.142} & \scalea{0.239} & \scalea{0.235} & \scalea{0.339} & \scalea{0.232} & \scalea{0.326} & \scalea{0.245} & \scalea{0.350} & \scalea{0.162} & \scalea{0.275} & \scalea{0.211} & \scalea{0.305} & \scalea{0.238} & \scalea{0.256} \\
    & \scalea{36}  & \scalea{\bst{0.140}} & \scalea{\bst{0.233}} & \scalea{\subbst{0.161}} & \scalea{\subbst{0.258}} & \scalea{0.182} & \scalea{0.282} & \scalea{0.202} & \scalea{0.305} & \scalea{0.167} & \scalea{0.258} & \scalea{0.197} & \scalea{0.300} & \scalea{0.379} & \scalea{0.428} & \scalea{0.344} & \scalea{0.417} & \scalea{0.234} & \scalea{0.331} & \scalea{0.250} & \scalea{0.338} & \scalea{0.263} & \scalea{0.277} \\
    & \scalea{48}  & \scalea{\bst{0.178}} & \scalea{\bst{0.265}} & \scalea{\subbst{0.206}} & \scalea{\subbst{0.293}} & \scalea{0.236} & \scalea{0.323} & \scalea{0.250} & \scalea{0.341} & \scalea{0.195} & \scalea{0.274} & \scalea{0.242} & \scalea{0.324} & \scalea{0.543} & \scalea{0.527} & \scalea{0.437} & \scalea{0.469} & \scalea{0.301} & \scalea{0.382} & \scalea{0.293} & \scalea{0.371} & \scalea{0.283} & \scalea{0.295} \\
    \cmidrule(lr){2-24}
    & \scalea{Avg}  & \scalea{\bst{0.127}} & \scalea{\bst{0.222}} & \scalea{\subbst{0.141}} & \scalea{\subbst{0.238}} & \scalea{0.159} & \scalea{0.258} & \scalea{0.174} & \scalea{0.277} & \scalea{0.154} & \scalea{0.245} & \scalea{0.717} & \scalea{0.459} & \scalea{0.319} & \scalea{0.378} & \scalea{0.294} & \scalea{0.377} & \scalea{0.201} & \scalea{0.303} & \scalea{0.232} & \scalea{0.322} & \scalea{0.249} & \scalea{0.266} \\
    \midrule

    \multicolumn{2}{c|}{\scalea{{$1^{\text{st}}$ Count}}} & \scalea{\bst{37}} & \scalea{\bst{40}} & \scalea{\subbst{2}} & \scalea{\subbst{2}} & \scalea{0} & \scalea{0} & \scalea{0} & \scalea{0} & \scalea{1} & \scalea{0} & \scalea{5} & \scalea{4} & \scalea{0} & \scalea{0} & \scalea{0} & \scalea{0} & \scalea{0} & \scalea{0} & \scalea{1} & \scalea{0} & \scalea{0} & \scalea{0} \\
    \bottomrule
  \end{tabular}%
  }
  \end{threeparttable}
  \vspace{-5pt}
\end{table}

\subsection{Overall Performance}\label{sec:overall}

Table~\ref{tab:longterm_app} reports the main long-term forecasting results across nine benchmarks and multiple prediction horizons. \method\ achieves the best average performance on most datasets and remains competitive on the remaining cases, with consistent gains over the iTransformer baseline on both MSE and MAE. The improvements are especially meaningful on high-dimensional benchmarks such as ECL, Traffic, PEMS03, and PEMS08, where modeling variables independently is more likely to distort cross-variable structure. These results suggest that CoRe can improve direct forecasting without changing the underlying architecture.

\begin{figure}
\begin{center}
\subfigure[ETTh2 snapshot.]{\includegraphics[width=0.31\linewidth]{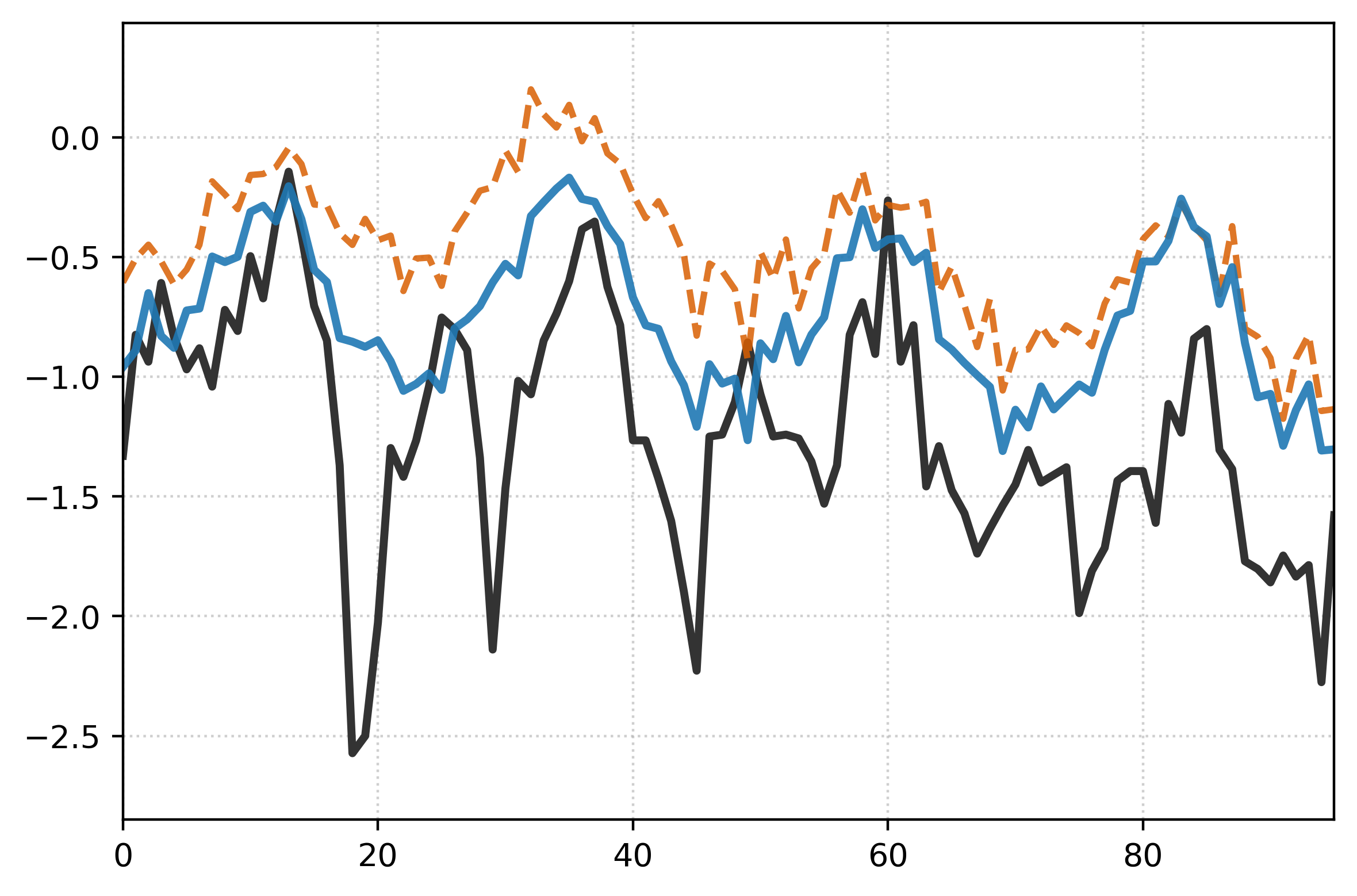}
}
\subfigure[Weather snapshot.]{
\includegraphics[width=0.31\linewidth]{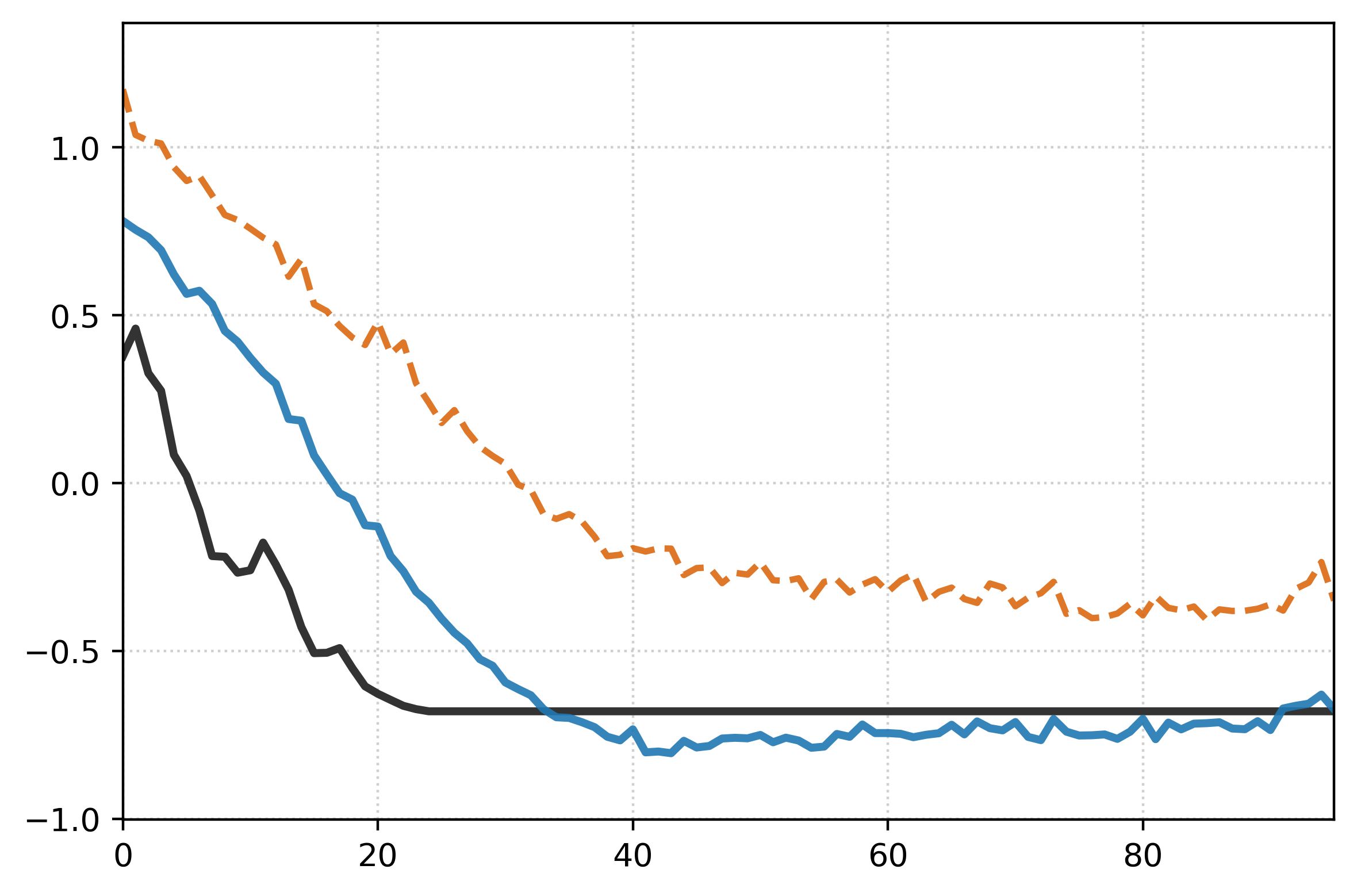}
}
\subfigure[ECL snapshot.]{
\includegraphics[width=0.31\linewidth]{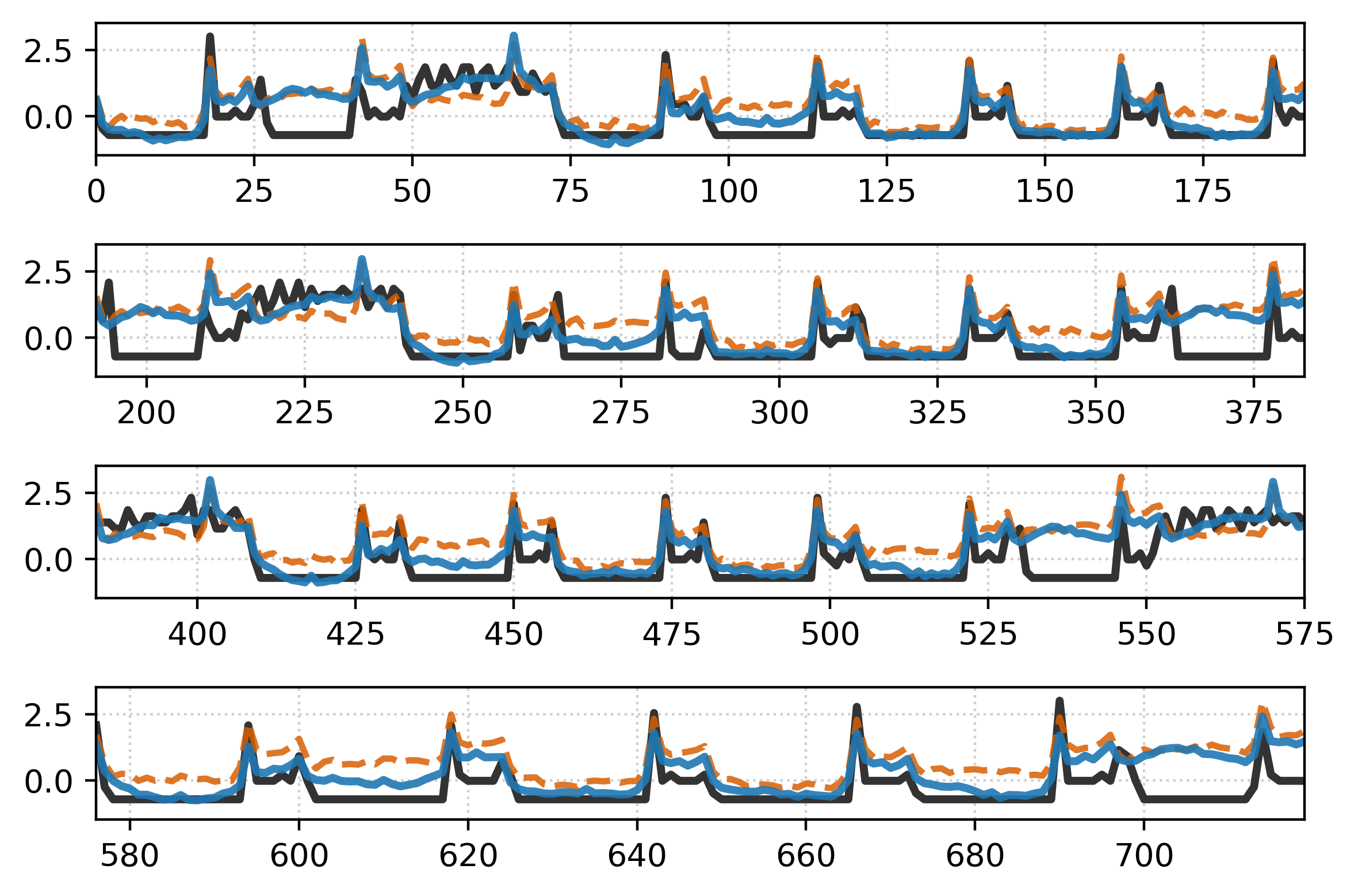}
}
\vspace{-5pt}
\caption{Forecast sequences of Ground Truth (black), DF (orange), and \method\ (blue), with historical length $\mathrm{H}=96$ in each case.}\label{fig:case}
\end{center}
\vspace{-10pt}
\end{figure}

\paragraph{Qualitative forecasts.} 
Fig.~\ref{fig:case} visualizes representative forecasts from ETTh2, Weather, and ECL. Standard DF captures the coarse trend in many cases, but it often smooths local fluctuations, underestimates sharp changes, or misses repeated peaks. In contrast, \method\ produces trajectories that stay closer to the ground truth, especially around short-term variations in ETTh2 and Weather and recurring peak patterns in ECL. These qualitative results are consistent with the design of CoRe: frequency coherence encourages global temporal alignment, while relational alignment discourages forecasts that fit each variable independently but distort the multivariate structure of future outputs.

\subsection{Ablation studies}\label{sec:ablation}

Table~\ref{tab:system_ablation_app} ablates the two components of CoRe. \method$^\dagger$ keeps the frequency coherence loss and removes the low-rank relational graph loss, while \method$^\ddagger$ keeps the relational graph loss and removes the frequency term. This design separates the temporal and cross-variable parts of the objective while leaving the backbone and training protocol unchanged. Both variants improve over the MSE baseline, indicating that temporal coherence and cross-variable relational alignment are individually useful for direct forecasting accuracy.

The graph-only variant is particularly effective on ETTh1, ECL, and short-horizon Weather, suggesting that preserving cross-variable relationships is important when the target variables are strongly coupled. The frequency-only variant remains competitive on ETTm1 and long-horizon Weather, where temporal structure and repeated patterns are prominent. Combining both components gives the strongest overall results, which supports the view that the two losses address different failure modes of pointwise supervision rather than acting as redundant regularizers during model training and evaluation.
\begin{table}
\caption{Ablation study results.}\label{tab:system_ablation_app}
\vspace{-5pt}
\setlength{\tabcolsep}{1pt}
\scriptsize
\centering
\begin{threeparttable}
\begin{tabular}{lcclccccccccccccccc}
    \toprule
    \multirow{2}{*}{Model} & \multirow{2}{*}{Fre.} & \multirow{2}{*}{Gra.} &\multirow{2}{*}{Data} && \multicolumn{2}{c}{T=96} && \multicolumn{2}{c}{T=192} && \multicolumn{2}{c}{T=336} && \multicolumn{2}{c}{T=720} && \multicolumn{2}{c}{Avg} \\
    \cmidrule{6-7} \cmidrule{9-10} \cmidrule{12-13} \cmidrule{15-16} \cmidrule{18-19}
    &&&&& MSE  & MAE && MSE & MAE && MSE & MAE && MSE & MAE && MSE & MAE \\
    \midrule
    
\multirow{4}{*}{MSE} & \multirow{4}{*}{\XSolidBrush} & \multirow{4}{*}{\XSolidBrush}
 &ETTm1&& 0.346 & 0.379 && 0.391 & 0.400 && 0.426 & 0.422 && 0.493 & 0.460 && 0.414 & 0.415 \\
 &&&ETTh1&& 0.390 & 0.409 && 0.442 & 0.440 && 0.479 & 0.457 && 0.483 & 0.479 && 0.449 & 0.446 \\
 &&&ECL&& 0.147 & 0.239 && 0.166 & 0.258 && 0.178 & 0.271 && 0.209 & 0.298 && 0.175 & 0.266 \\
 &&&Weather&& 0.201 & 0.246 && 0.250 & 0.282 && 0.302 & 0.317 && 0.370 & 0.361 && 0.280 & 0.302 \\
\midrule

\multirow{4}{*}{{\method}$^\dagger$} & \multirow{4}{*}{\Checkmark} & \multirow{4}{*}{\XSolidBrush}
 &ETTm1&& \subbst{0.324} & \subbst{0.362} && \subbst{0.372} & \subbst{0.385} && \subbst{0.402} & \subbst{0.404} && \subbst{0.468} & \subbst{0.443} && \subbst{0.391} & \subbst{0.398} \\
 &&&ETTh1&& 0.381 & 0.400 && 0.430 & 0.426 && 0.474 & 0.451 && 0.470 & 0.465 && 0.439 & 0.436\\
 &&&ECL&& \subbst{0.144} & \subbst{0.233} && 0.160 & 0.250 && 0.174 & 0.265 && \subbst{0.204} & \subbst{0.293} && 0.171 & \subbst{0.260}\\
 &&&Weather&& 0.166 & 0.205 && 0.220 & 0.252 && \subbst{0.274} & \subbst{0.293} && \subbst{0.356} & \subbst{0.346} && \subbst{0.254} & \subbst{0.274}\\
\midrule

\multirow{4}{*}{{\method}$^\ddagger$} & \multirow{4}{*}{\XSolidBrush} & \multirow{4}{*}{\Checkmark}
 &ETTm1&& 0.335 & 0.370 && 0.382 & 0.392 && 0.415 & 0.412 && 0.480 & 0.452 && 0.403 & 0.406 \\
 &&&ETTh1&& \subbst{0.380} & \subbst{0.398} && \subbst{0.429} & \subbst{0.424} && \subbst{0.468} & \subbst{0.448} && \subbst{0.459} & \subbst{0.455} && \subbst{0.434} & \subbst{0.431} \\
 &&&ECL&& 0.145 & 0.236 && \subbst{0.158} & \subbst{0.248} && \subbst{0.172} & \subbst{0.263} && 0.206 & 0.295 && \subbst{0.170} & 0.261 \\
 &&&Weather&& \subbst{0.164} & \subbst{0.203} && \subbst{0.219} & \subbst{0.251} && 0.285 & 0.300 && 0.360 & 0.350 && 0.257 & 0.278 \\
\midrule

\multirow{4}{*}{{\method}} & \multirow{4}{*}{\Checkmark} & \multirow{4}{*}{\Checkmark}
 &ETTm1&& \bst{0.308} & \bst{0.346} && \bst{0.360} & \bst{0.373} && \bst{0.393} & \bst{0.395} && \bst{0.462} & \bst{0.434} && \bst{0.381} & \bst{0.387} \\
 &&&ETTh1&& \bst{0.376} & \bst{0.391} && \bst{0.424} & \bst{0.425} && \bst{0.463} & \bst{0.446} && \bst{0.450} & \bst{0.449} && \bst{0.428} & \bst{0.428}\\
 &&&ECL&& \bst{0.143} & \bst{0.231} && \bst{0.156} & \bst{0.246} && \bst{0.169} & \bst{0.259} && \bst{0.202} & \bst{0.289} && \bst{0.168} & \bst{0.256}\\
 &&&Weather&& \bst{0.161} & \bst{0.200} && \bst{0.216} & \bst{0.250} && \bst{0.272} & \bst{0.291} && \bst{0.353} & \bst{0.345} && \bst{0.251} & \bst{0.272}\\
    \bottomrule
\end{tabular}
\begin{tablenotes}
    \item  \tiny \textit{Note}:  \bst{Bold} and \subbst{underlined} denote best and second-best results, respectively. “Fre.” and “Gra.” are abbreviations for frequency and graph.
\end{tablenotes}
\end{threeparttable}
\vspace{-5pt}
\end{table}

\begin{table}[h]
  \caption{Comparable results with other objectives for time-series forecast.}\label{tab:loss_avg}
  \vspace{-5pt}
  \renewcommand{\arraystretch}{1.2} 
  \setlength{\tabcolsep}{3pt}
  \scriptsize
  \centering
  \renewcommand{\multirowsetup}{\centering}
  \begin{threeparttable}
    \resizebox{\columnwidth}{!}{
      \begin{tabular}{c|l|cc|cc|cc|cc|cc|cc|cc}
        \toprule
        \multicolumn{2}{l}{Loss} & 
        \multicolumn{2}{c}{\textbf{CoRe}} &
        \multicolumn{2}{c}{Time-o1} &
        \multicolumn{2}{c}{FreDF} &
        \multicolumn{2}{c}{Koopman} &
        \multicolumn{2}{c}{Dilate} &
        \multicolumn{2}{c}{Soft-DTW} &
        \multicolumn{2}{c}{MSE} \\
        \cmidrule(lr){3-4} \cmidrule(lr){5-6} \cmidrule(lr){7-8} \cmidrule(lr){9-10} \cmidrule(lr){11-12} \cmidrule(lr){13-14} \cmidrule(lr){15-16}
        \multicolumn{2}{l}{Metrics}  & MSE & MAE  & MSE & MAE  & MSE & MAE  & MSE & MAE  & MSE & MAE  & MSE & MAE  & MSE & MAE  \\
        \toprule
        \multirow{4}{*}{{\rotatebox{90}{\scaleb{iTransformer}}}}
        & ETTm1 & \bst{0.381} & \bst{0.387} & \subbst{0.395} & \subbst{0.401} & 0.405 & 0.405 & 0.413 & 0.416 & 0.407 & 0.412 & 0.417 & 0.415 & 0.411 & 0.414 \\
        & ETTh1 & \bst{0.428} & \bst{0.428} & \subbst{0.438} & \subbst{0.434} & 0.442 & 0.437 & 0.455 & 0.451 & 0.452 & 0.448 & 0.470 & 0.457 & 0.452 & 0.448 \\
        & ECL & \bst{0.168} & \bst{0.256} & \subbst{0.170} & \subbst{0.260} & 0.176 & 0.264 & 0.178 & 0.269 & 0.178 & 0.269 & 0.175 & 0.266 & 0.179 & 0.270 \\
        & Weather & \bst{0.251} & \bst{0.272} & \bst{0.251} & \bst{0.272} & 0.257 & 0.276 & 0.289 & 0.313 & 0.286 & 0.309 & 0.292 & 0.316 & 0.269 & 0.289 \\
        \bottomrule
      \end{tabular}
    }
\begin{tablenotes}
    \item  \scriptsize \textit{Note}:  \bst{Bold} and \subbst{underlined} denote best and second results. Follow the settings of Time-o1~\cite{wang2025nipstimeo1}
\end{tablenotes}
  \end{threeparttable}
\end{table}

\subsection{Learning objective comparison}
\label{sec:compete}

Table~\ref{tab:loss_avg} compares \method\ with recent time-series learning objectives, including Time-o1~\cite{wang2025nipstimeo1}, FreDF~\cite{wang2025fredf}, Koopman~\cite{koopman}, DILATE~\cite{Dilate}, and Soft-DTW~\cite{soft-dtw}. For a fair comparison, we integrate their official implementations into iTransformer~\cite{itransformer} and keep the same backbone, data splits, and evaluation metrics. This setting isolates the effect of the objective design from architectural differences.

As shown in Table~\ref{tab:loss_avg}, \method\ achieves the best or tied-best results across the four datasets. It improves over the MSE baseline on both MSE and MAE, supporting the benefit of replacing purely pointwise supervision with structural output constraints. Compared with FreDF and Time-o1, CoRe remains competitive on Weather and yields clearer gains on ETTm1, ETTh1, and ECL. These results suggest that cross-variable relational consistency provides information complementary to temporal objective design, especially in multivariate settings where variables should not be optimized independently.

\subsection{Generalization Studies}\label{sec:generalize}

\begin{figure*}[t]
\begin{center}
\subfigure[\scriptsize ETTm1 with MSE]{
    \includegraphics[width=0.22\linewidth]{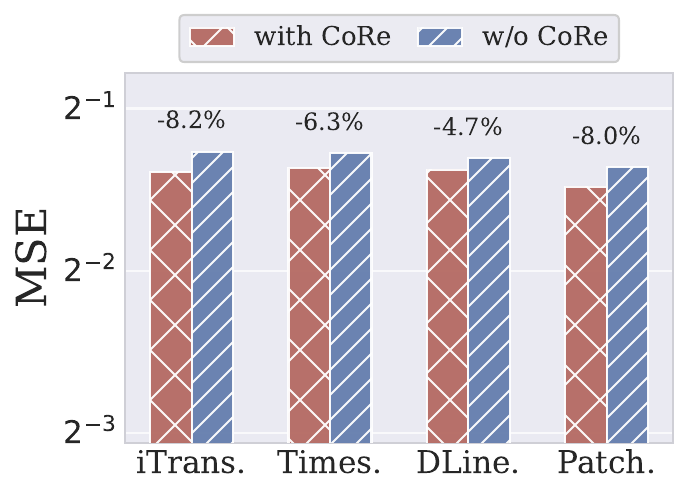}
}
\subfigure[\scriptsize ETTm1 with MAE]{
    \includegraphics[width=0.22\linewidth]{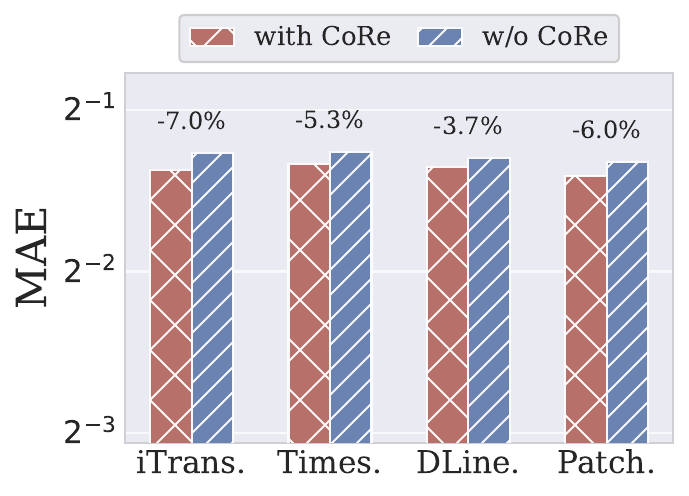}
}
\subfigure[\scriptsize ETTm2 with MSE]{
    \includegraphics[width=0.22\linewidth]{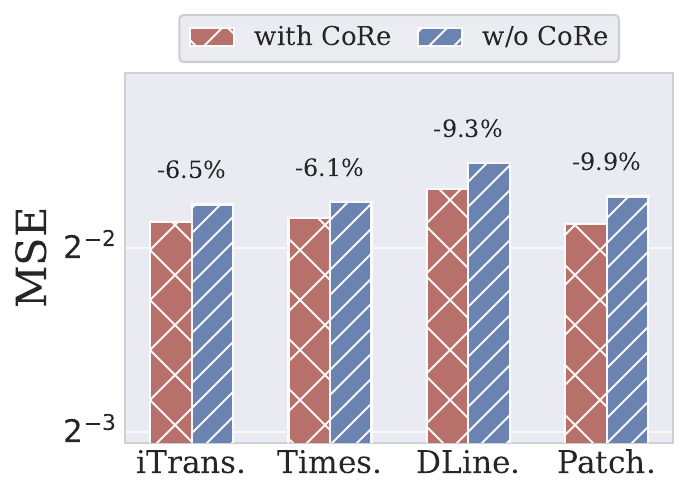}
}
\subfigure[\scriptsize ETTm2 with MAE]{
    \includegraphics[width=0.22\linewidth]{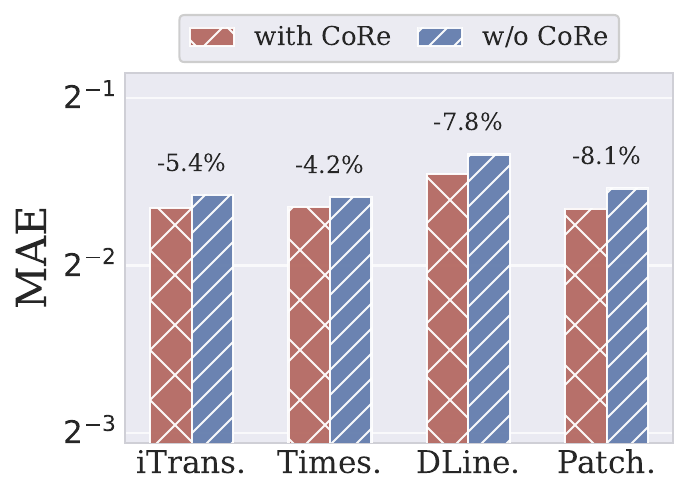}
}

\subfigure[\scriptsize ETTh1 with MSE]{
    \includegraphics[width=0.22\linewidth]{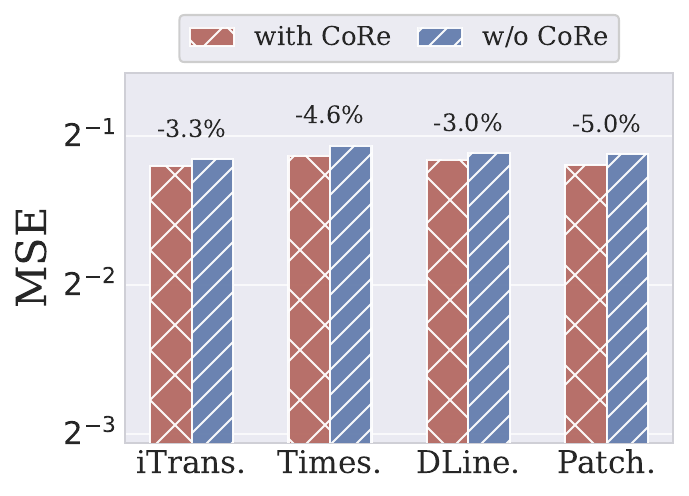}
}
\subfigure[\scriptsize ETTh1 with MAE]{
    \includegraphics[width=0.22\linewidth]{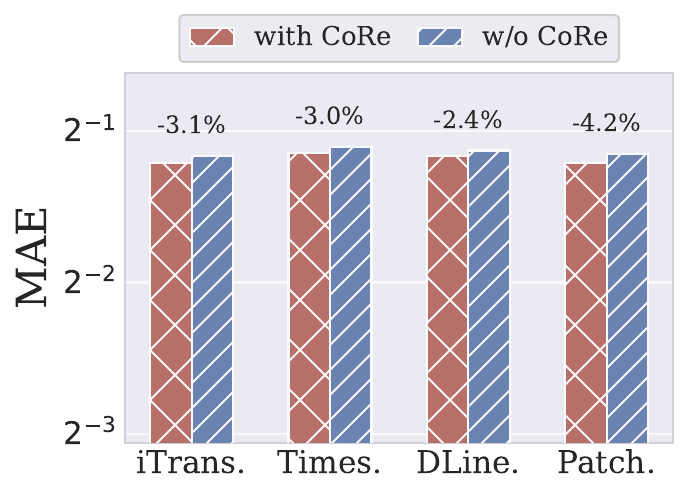}
}
\subfigure[\scriptsize ETTh2 with MSE]{
    \includegraphics[width=0.22\linewidth]{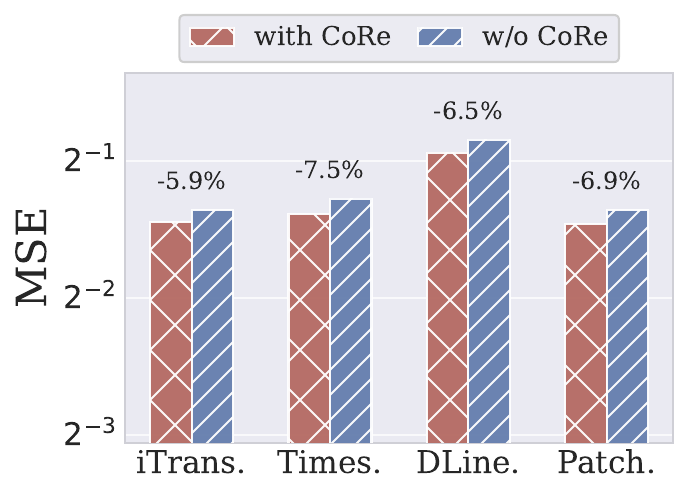}
}
\subfigure[\scriptsize ETTh2 with MAE]{
    \includegraphics[width=0.22\linewidth]{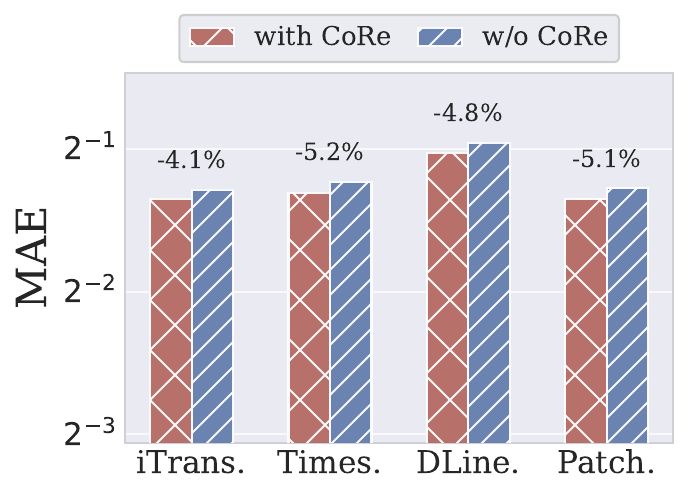}
}

\subfigure[\scriptsize ECL with MSE]{
    \includegraphics[width=0.22\linewidth]{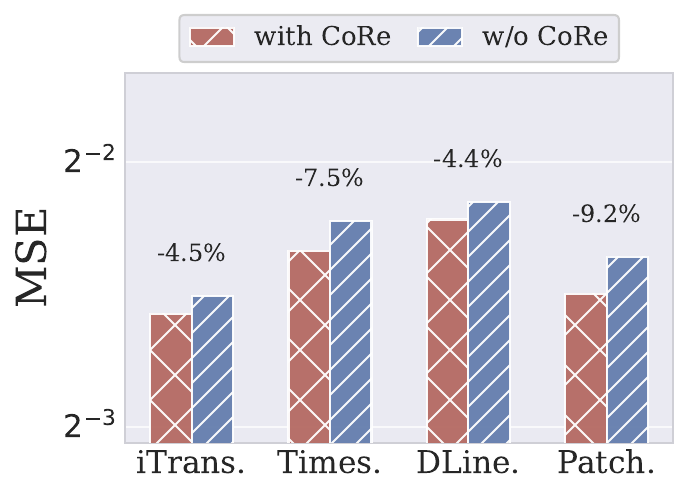}
}
\subfigure[\scriptsize ECL with MAE]{
    \includegraphics[width=0.22\linewidth]{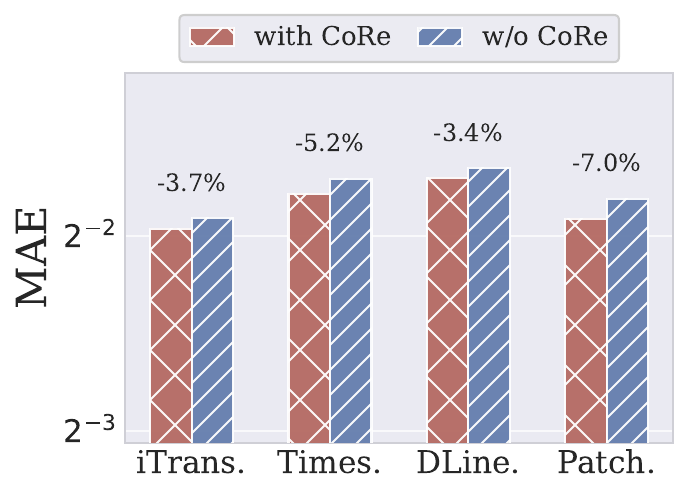}
}
\subfigure[\scriptsize Weather with MSE]{
    \includegraphics[width=0.22\linewidth]{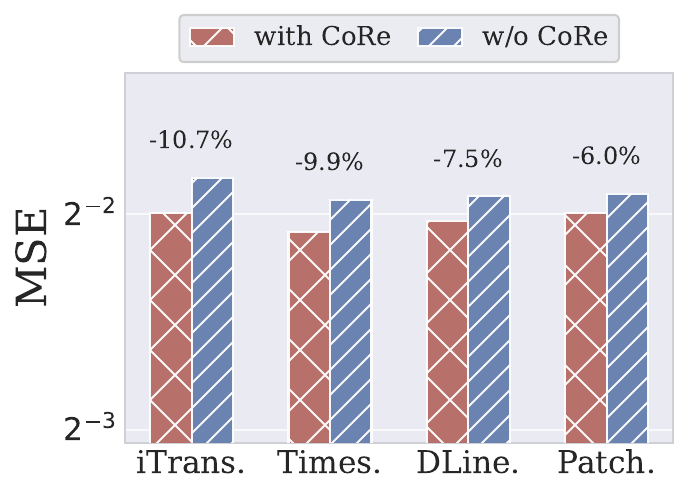}
}
\subfigure[\scriptsize Weather with MAE]{
    \includegraphics[width=0.22\linewidth]{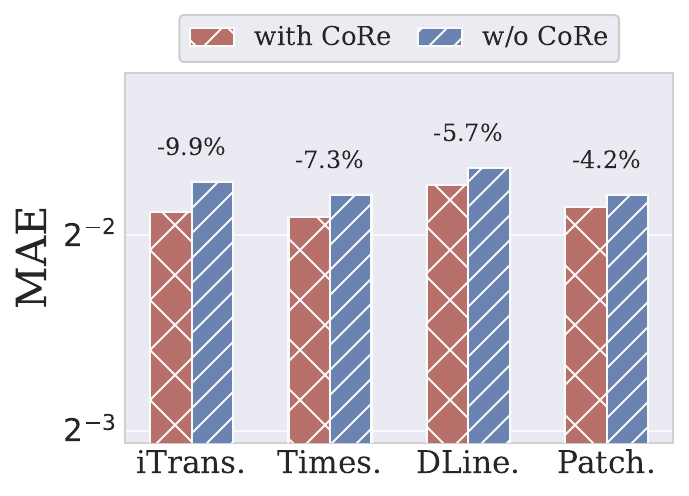}
}
\vspace{-5pt}
\caption{Effect of incorporating \method\ into different forecasting backbones across six datasets in our experiments and evaluation protocol.}
\label{fig:backbone_all}
\end{center}
\end{figure*}

\paragraph{Varying forecast models.}
Figure~\ref{fig:backbone_all} integrates CoRe with iTransformer~\cite{itransformer}, TimesNet~\cite{Timesnet}, DLinear~\cite{DLinear}, and PatchTST~\cite{PatchTST}, covering Transformer, convolutional, linear, and patch-based forecasting designs. CoRe improves both MSE and MAE for all reported backbone--dataset combinations, with larger gains on challenging datasets such as Weather, ETTm2, and ECL. The consistency across these heterogeneous models indicates that CoRe is not compensating for a particular architectural weakness. Instead, it provides a model-agnostic training signal that can be attached to different direct forecasting backbones.

\paragraph{Varying relational graph loss variants.}\label{sec:variants} 
\looseness=-1 We compare three relational graph losses, each combined with the same frequency coherence loss and the same training protocol.

\begin{itemize}[leftmargin=*, label=\textbullet]
    \item \textbf{LR-Diff.}
    This is the default relational loss used in CoRe. It projects predictions and targets into the target-derived PCA subspace and matches first-order differences between randomly sampled latent component pairs.
    
    \item \textbf{Cov.}
    It aligns prediction and ground-truth covariance matrices in the original variable space after separate centering, providing a global constraint on cross-variable dependency among the observed variables.
    
    \item \textbf{PCA-Cov.}
    It uses the same target-derived PCA projection as LR-Diff, but replaces sampled pairwise difference matching with covariance alignment in the low-rank subspace of latent variables.
\end{itemize}

As shown in Table~\ref{tab:graph_variant_comparison}, LR-Diff performs best in most cases. Cov and PCA-Cov are also competitive, showing that covariance-based relational alignment can improve multivariate forecasting. However, they are generally weaker than LR-Diff, possibly because global covariance statistics discard fine-grained pairwise differences that are directly relevant to preserving relative variable behavior. We therefore use LR-Diff as the default relational graph loss in {\method}.

\begin{table}[h]
\centering
\caption{Comparison of different graph loss variants.}
\label{tab:graph_variant_comparison}
\vspace{-5pt}
\setlength{\tabcolsep}{3.5pt}
\resizebox{0.9\textwidth}{!}{%
\begin{tabular}{clcccccccccc}
\toprule
\multirow{2}{*}{Data} & \multirow{2}{*}{Variant} 
& \multicolumn{2}{c}{T=96} 
& \multicolumn{2}{c}{T=192} 
& \multicolumn{2}{c}{T=336} 
& \multicolumn{2}{c}{T=720} 
& \multicolumn{2}{c}{Avg} \\
\cmidrule(lr){3-4} \cmidrule(lr){5-6} \cmidrule(lr){7-8} \cmidrule(lr){9-10} \cmidrule(lr){11-12}
& & MSE & MAE & MSE & MAE & MSE & MAE & MSE & MAE & MSE & MAE \\
\midrule
\multirow{3}{*}{\rotatebox{90}{\scalebox{0.95}{ETTm1}}} 
& Diff    & \bst{0.308} & \bst{0.346} & \bst{0.360} & \bst{0.373} & \bst{0.393} & \bst{0.395} & \bst{0.462} & \bst{0.434} & \bst{0.381} & \bst{0.387} \\
& Cov     & \subbst{0.325} & \subbst{0.360} & \subbst{0.373} & \subbst{0.385} & \subbst{0.404} & \subbst{0.405} & \subbst{0.468} & \subbst{0.442} & \subbst{0.392} & \subbst{0.398} \\
& PCA-Cov & 0.326 & 0.361 & \subbst{0.373} & \subbst{0.385} & \subbst{0.404} & \subbst{0.405} & 0.468 & 0.442 & 0.393 & 0.398 \\
\midrule
\multirow{3}{*}{\rotatebox{90}{\scalebox{0.95}{ETTm2}}} 
& Diff    & \bst{0.172} & \bst{0.250} & \bst{0.237} & \bst{0.294} & \bst{0.296} & \bst{0.333} & \bst{0.394} & \bst{0.391} & \bst{0.275} & \bst{0.317} \\
& Cov     & \subbst{0.173} & \subbst{0.251} & 0.239 & 0.296 & 0.298 & 0.334 & 0.396 & 0.391 & 0.276 & 0.318 \\
& PCA-Cov & 0.174 & 0.252 & \subbst{0.238} & \subbst{0.295} & \subbst{0.296} & \subbst{0.333} & \subbst{0.395} & \subbst{0.391} & \subbst{0.276} & \subbst{0.318} \\
\midrule
\multirow{3}{*}{\rotatebox{90}{\scalebox{0.95}{ETTh1}}} 
& Diff    & \bst{0.376} & \bst{0.391} & \bst{0.424} & \bst{0.425} & \bst{0.463} & \bst{0.446} & \bst{0.473} & \bst{0.471} & \bst{0.434} & \bst{0.433} \\
& Cov     & \subbst{0.382} & \subbst{0.396} & \subbst{0.426} & \subbst{0.427} & 0.472 & \subbst{0.446} & \subbst{0.492} & 0.477 & \subbst{0.443} & \subbst{0.436} \\
& PCA-Cov & \subbst{0.382} & \subbst{0.396} & 0.426 & 0.427 & \subbst{0.469} & 0.448 & 0.495 & \subbst{0.476} & 0.443 & 0.437 \\
\midrule
\multirow{3}{*}{\rotatebox{90}{\scalebox{0.95}{ETTh2}}} 
& Diff    & \bst{0.285} & \bst{0.334} & \bst{0.363} & \bst{0.383} & \bst{0.409} & \bst{0.420} & \bst{0.413} & \bst{0.435} & \bst{0.367} & \bst{0.393} \\
& Cov     & 0.285 & 0.335 & 0.364 & \subbst{0.384} & \subbst{0.412} & \subbst{0.421} & 0.417 & 0.437 & 0.369 & 0.394 \\
& PCA-Cov & \subbst{0.285} & \subbst{0.335} & \subbst{0.363} & 0.384 & \subbst{0.412} & \subbst{0.421} & \subbst{0.416} & \subbst{0.437} & \subbst{0.369} & \subbst{0.394} \\
\midrule
\multirow{3}{*}{\rotatebox{90}{\scalebox{0.95}{ECL}}} 
& Diff    & \bst{0.143} & \bst{0.231} & \bst{0.158} & \bst{0.246} & \bst{0.171} & \bst{0.261} & \bst{0.202} & \bst{0.289} & \bst{0.168} & \bst{0.257} \\
& Cov     & 0.144 & 0.233 & 0.159 & \subbst{0.248} & \subbst{0.172} & \subbst{0.263} & \subbst{0.205} & \subbst{0.291} & \subbst{0.170} & \subbst{0.259} \\
& PCA-Cov & \subbst{0.144} & \subbst{0.233} & \subbst{0.159} & 0.248 & 0.172 & 0.263 & 0.205 & 0.293 & 0.170 & 0.259 \\
\midrule
\multirow{3}{*}{\rotatebox{90}{\scalebox{0.95}{Weather}}} 
& Diff    & \bst{0.163} & \bst{0.202} & \bst{0.216} & \bst{0.250} & \bst{0.275} & \bst{0.293} & \bst{0.353} & \bst{0.345} & \bst{0.252} & \bst{0.272} \\
& Cov     & \subbst{0.163} & \subbst{0.202} & \subbst{0.217} & \subbst{0.251} & 0.277 & 0.294 & 0.354 & \subbst{0.346} & 0.253 & \subbst{0.273} \\
& PCA-Cov & 0.165 & 0.203 & 0.217 & 0.252 & \subbst{0.275} & \subbst{0.294} & \subbst{0.354} & 0.346 & \subbst{0.253} & 0.274 \\
\bottomrule
\end{tabular}%
}
\end{table}

\subsection{Hyperparameter Sensitivity}\label{sec:hyper}

\begin{figure}[!htbp]
\begin{center}
\subfigure[\scriptsize ECL MSE]{\includegraphics[width=0.24\linewidth]{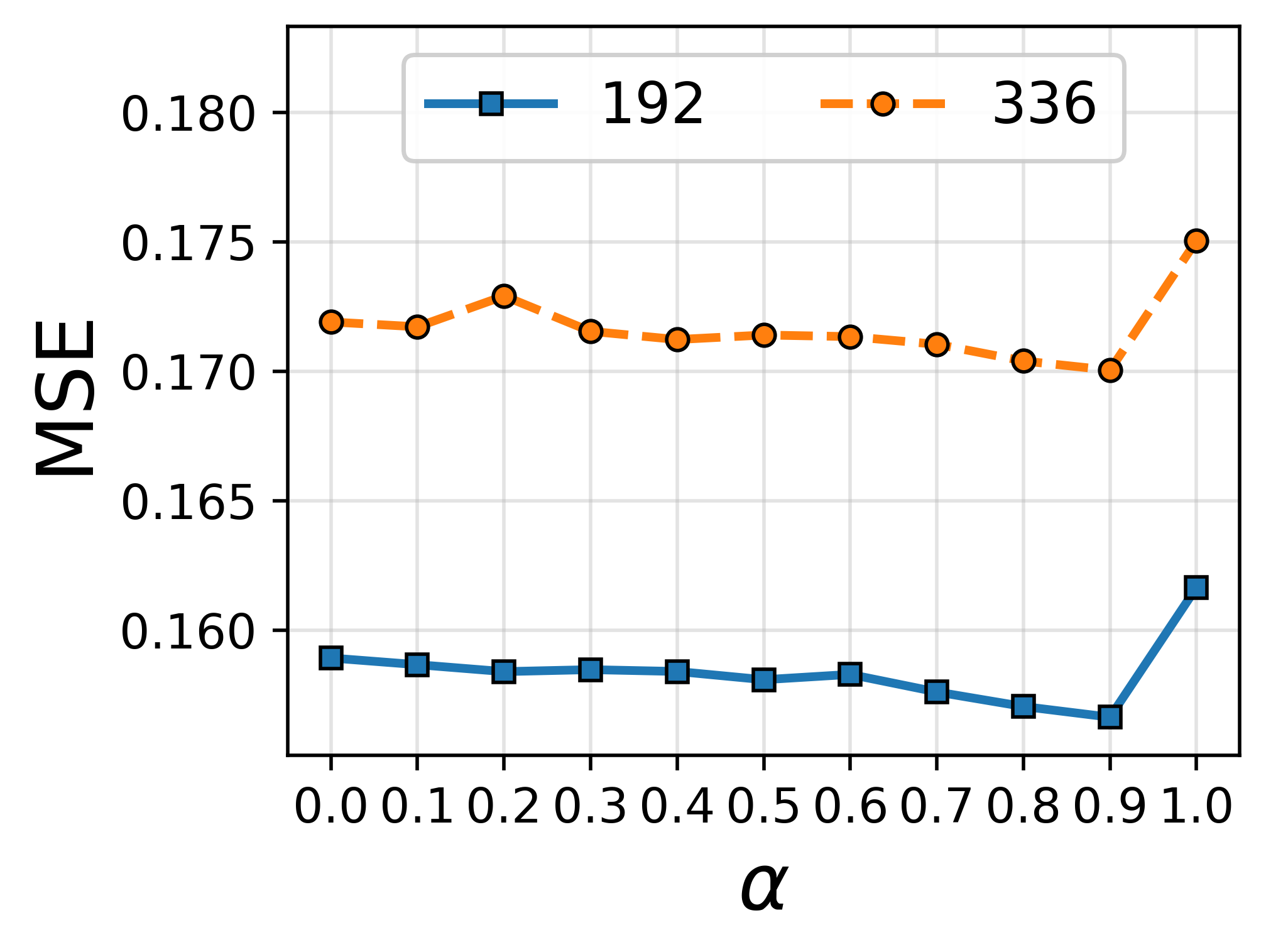}}
\subfigure[\scriptsize ECL MAE]{\includegraphics[width=0.24\linewidth]{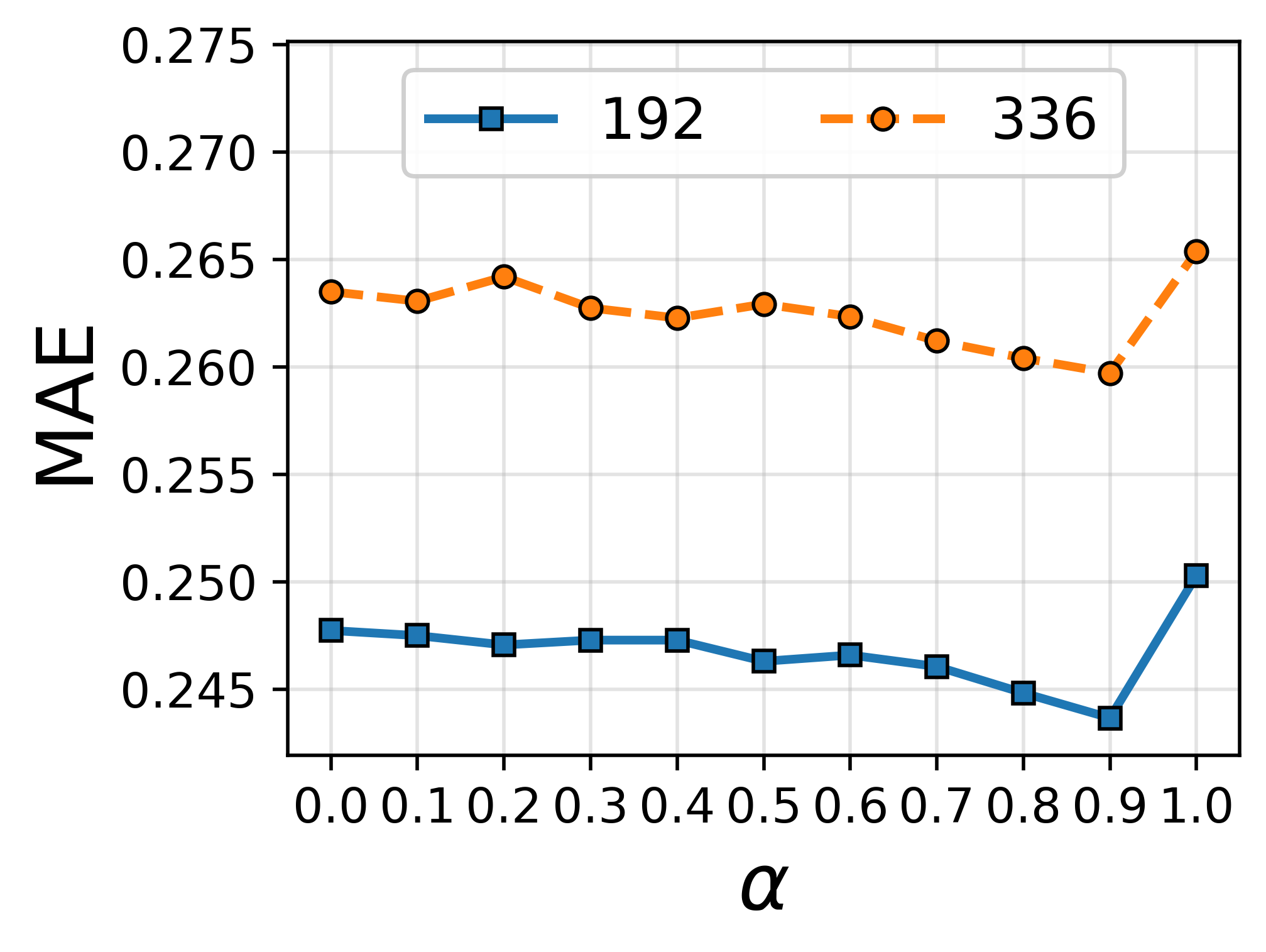}}
\subfigure[\scriptsize ETTm1 MSE]{\includegraphics[width=0.24\linewidth]{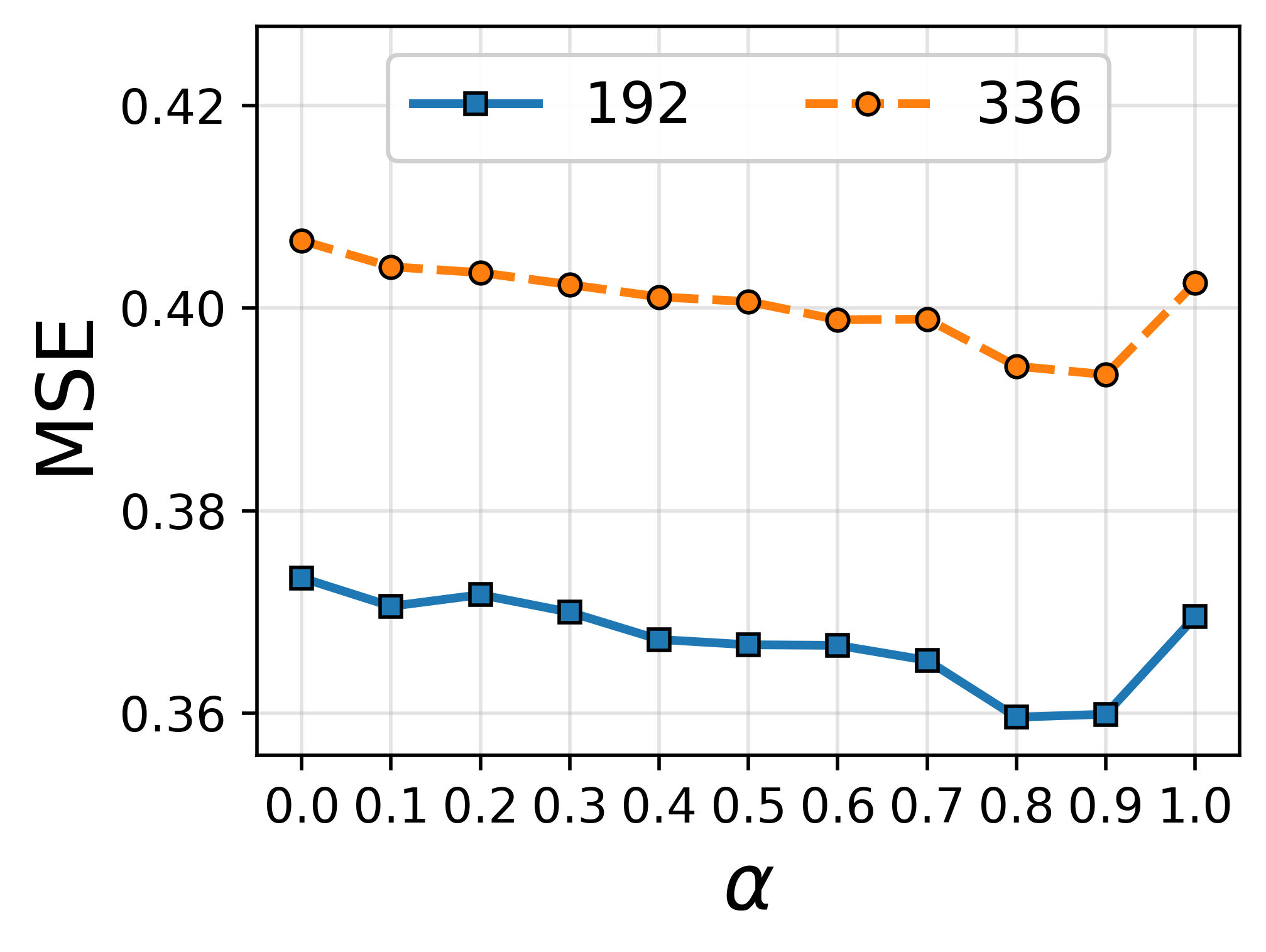}}
\subfigure[\scriptsize ETTm1 MAE]{\includegraphics[width=0.24\linewidth]{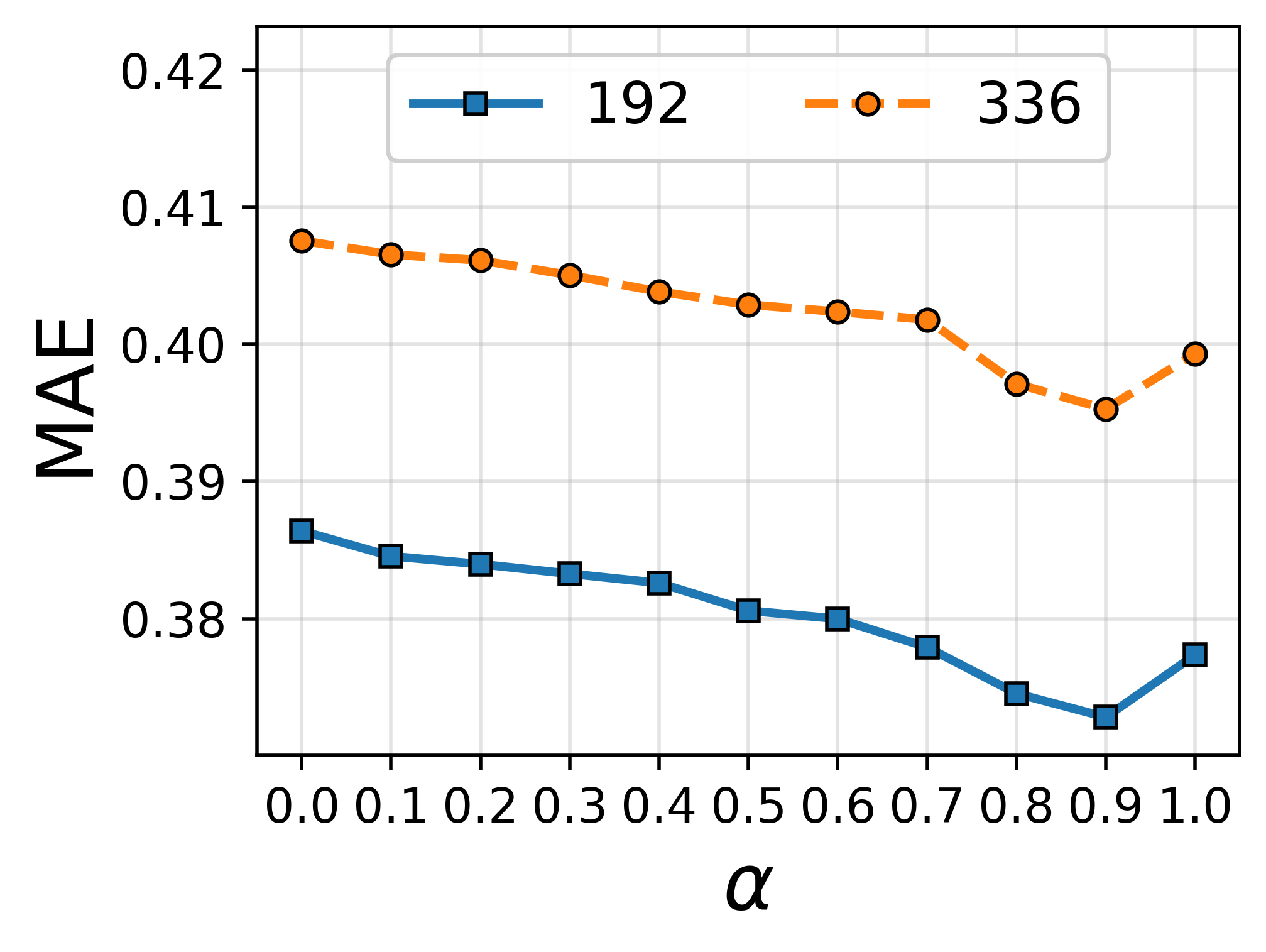}}
\vspace{-0.5em}

\subfigure[\scriptsize Weather MSE]{\includegraphics[width=0.24\linewidth]{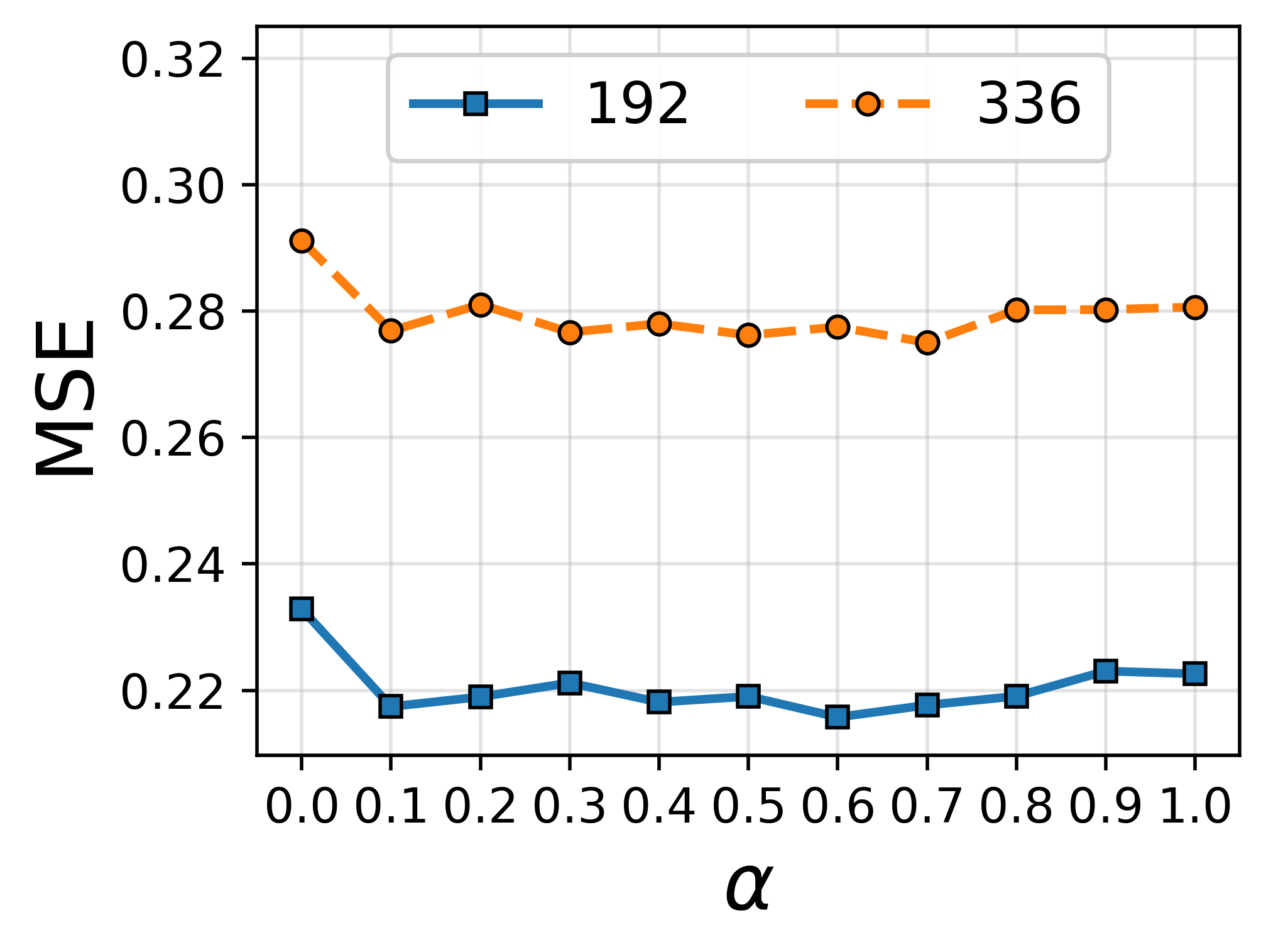}}
\subfigure[\scriptsize Weather MAE]{\includegraphics[width=0.24\linewidth]{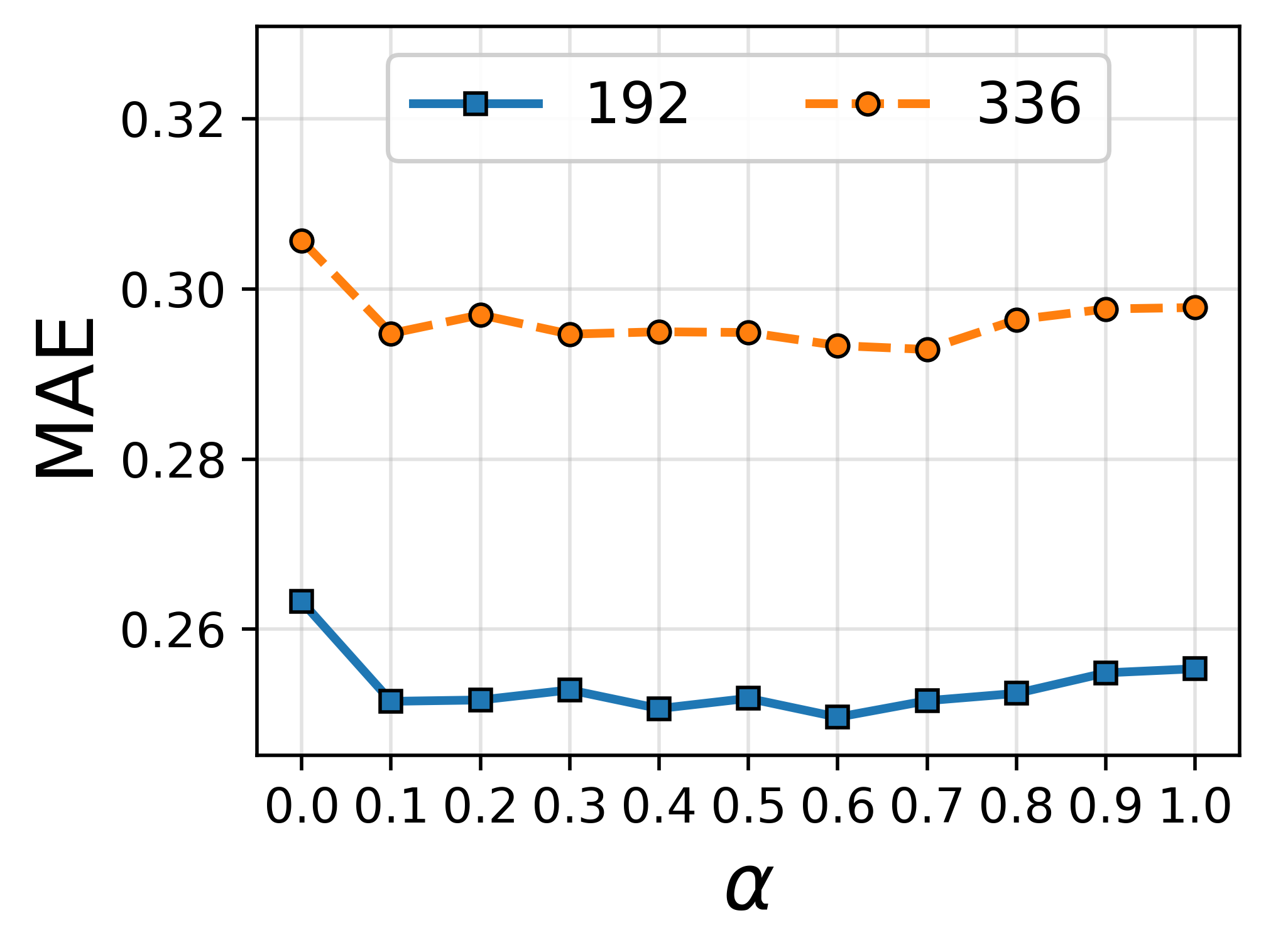}}
\subfigure[\scriptsize ETTh2 MSE]{\includegraphics[width=0.24\linewidth]{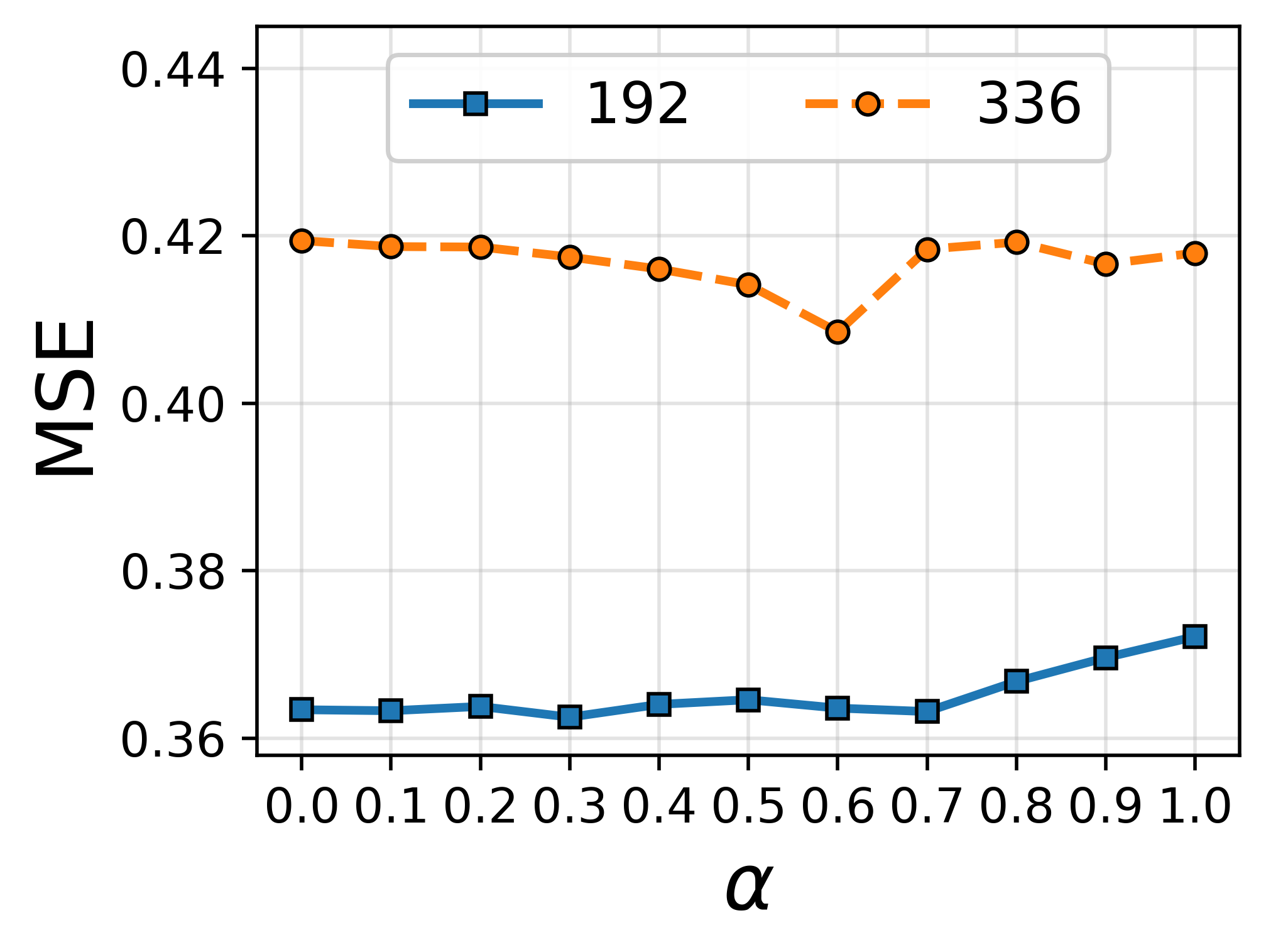}}
\subfigure[\scriptsize ETTh2 MAE]{\includegraphics[width=0.24\linewidth]{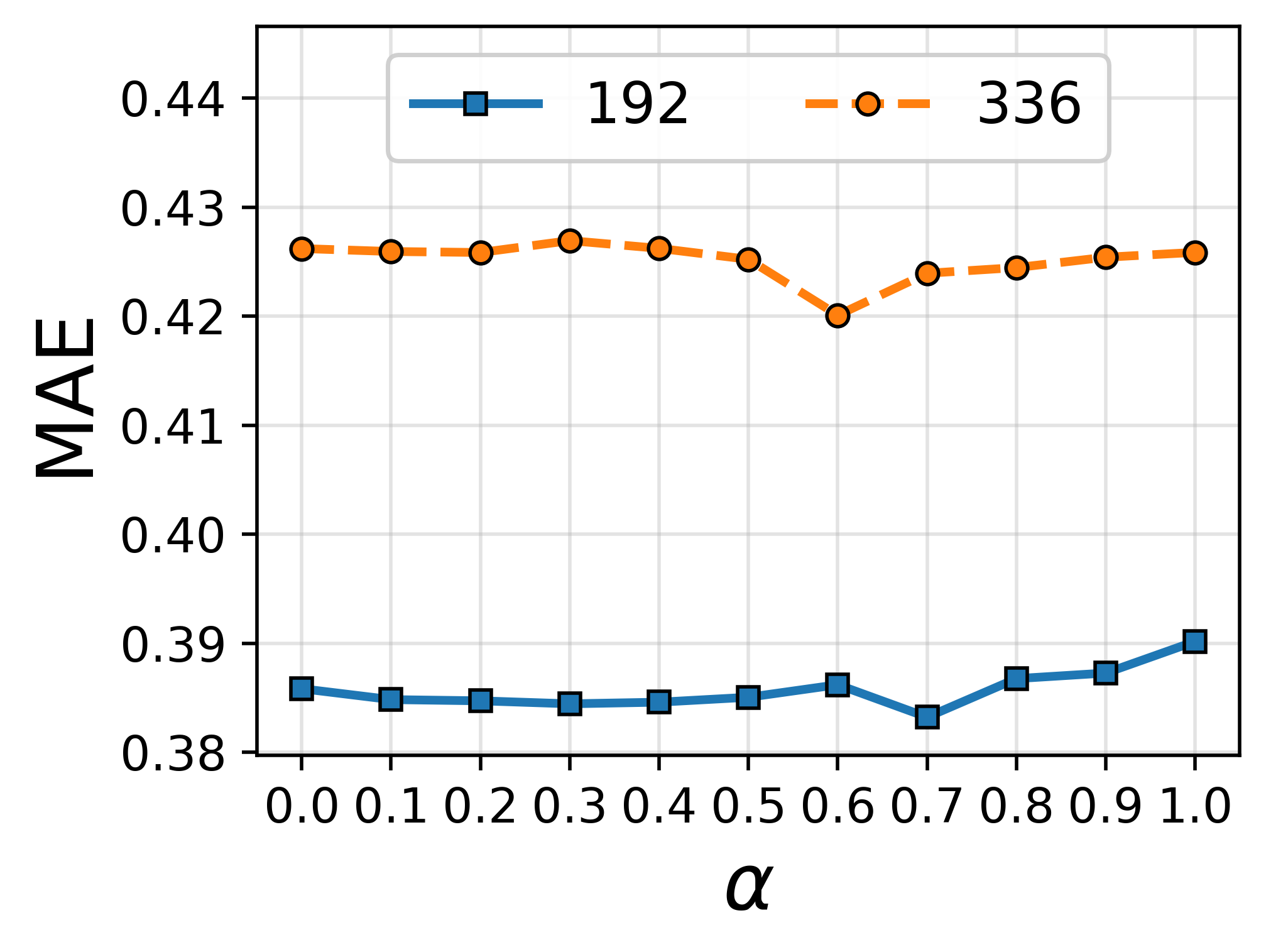}}
\vspace{-0.8em}
\caption{Sensitivity to the loss weight $\alpha$ for prediction lengths $T=192$ and $T=336$.}
\label{fig:sensi}
\end{center}
\vspace{-1.4em}
\end{figure}

\paragraph{Sensitivity of loss weight $\alpha$.}\label{sec:sensi_alpha} 
We vary the loss weight $\alpha$ from 0 to 1 on ECL, ETTm1, Weather, and ETTh2 with prediction lengths $T=192$ and $T=336$. As shown in Fig.~\ref{fig:sensi}, MSE and MAE change smoothly in most cases, without abrupt degradation when $\alpha$ moves away from the best value. This indicates that CoRe does not require delicate balancing between the frequency coherence loss and the low-rank relational graph loss. In practice, a broad range of intermediate weights provides stable gains, which is important when transferring the objective to new forecasting datasets and backbones.

\begin{figure}[!htbp]
\centering
\subfigure[\scriptsize ETTh1]{\includegraphics[width=0.24\linewidth]{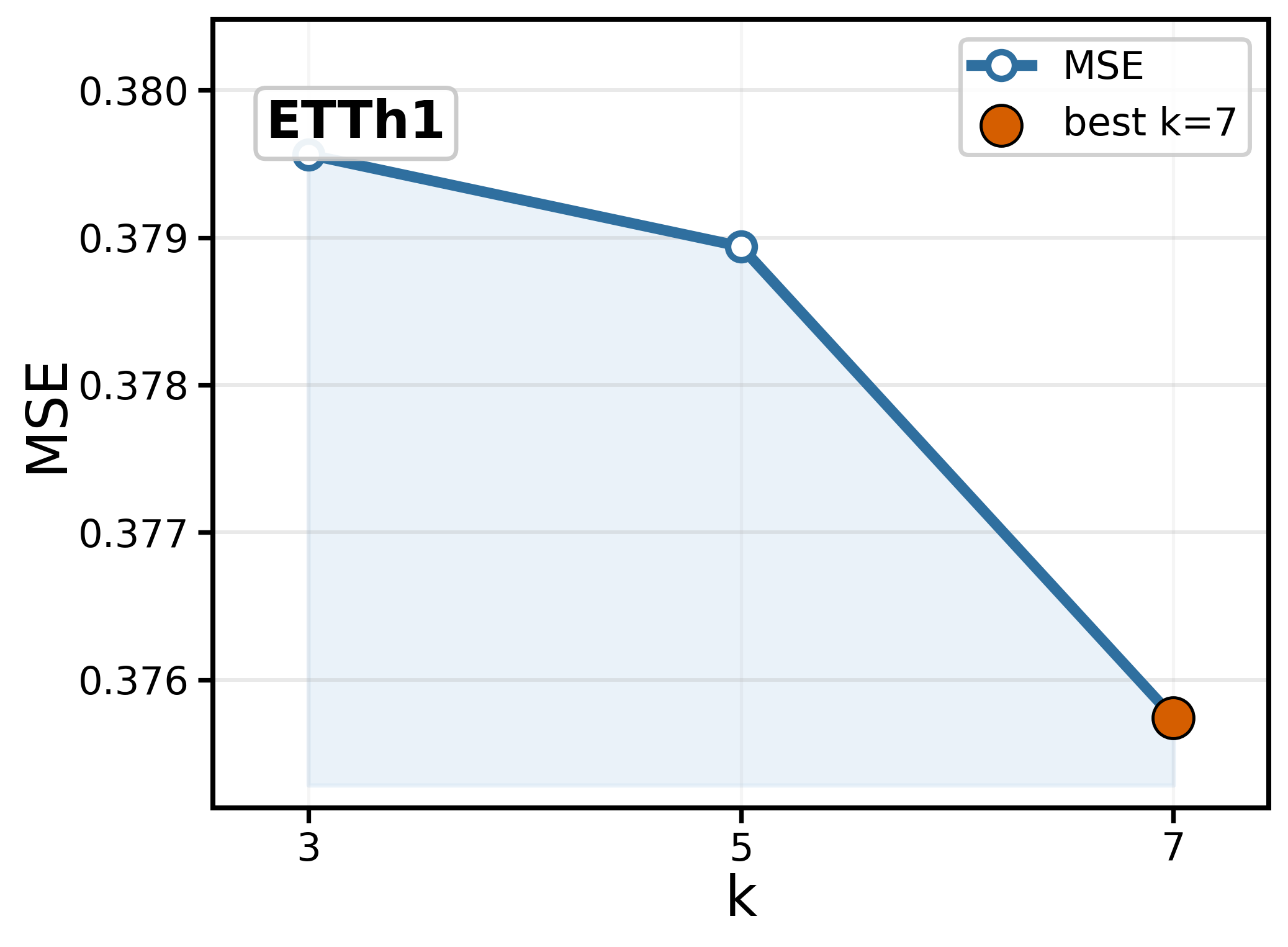}}
\subfigure[\scriptsize ETTh2]{\includegraphics[width=0.24\linewidth]{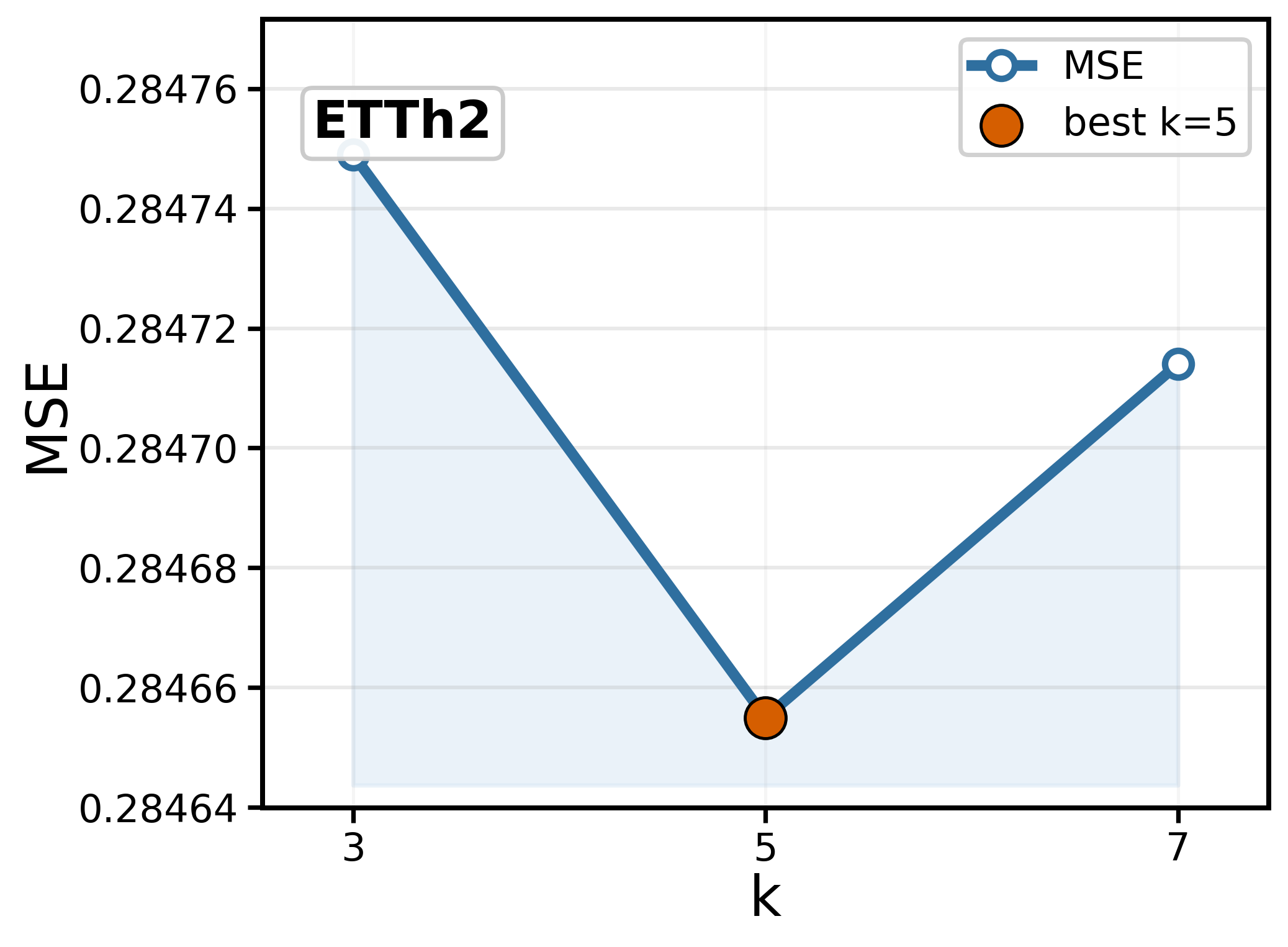}}
\subfigure[\scriptsize ETTm1]{\includegraphics[width=0.24\linewidth]{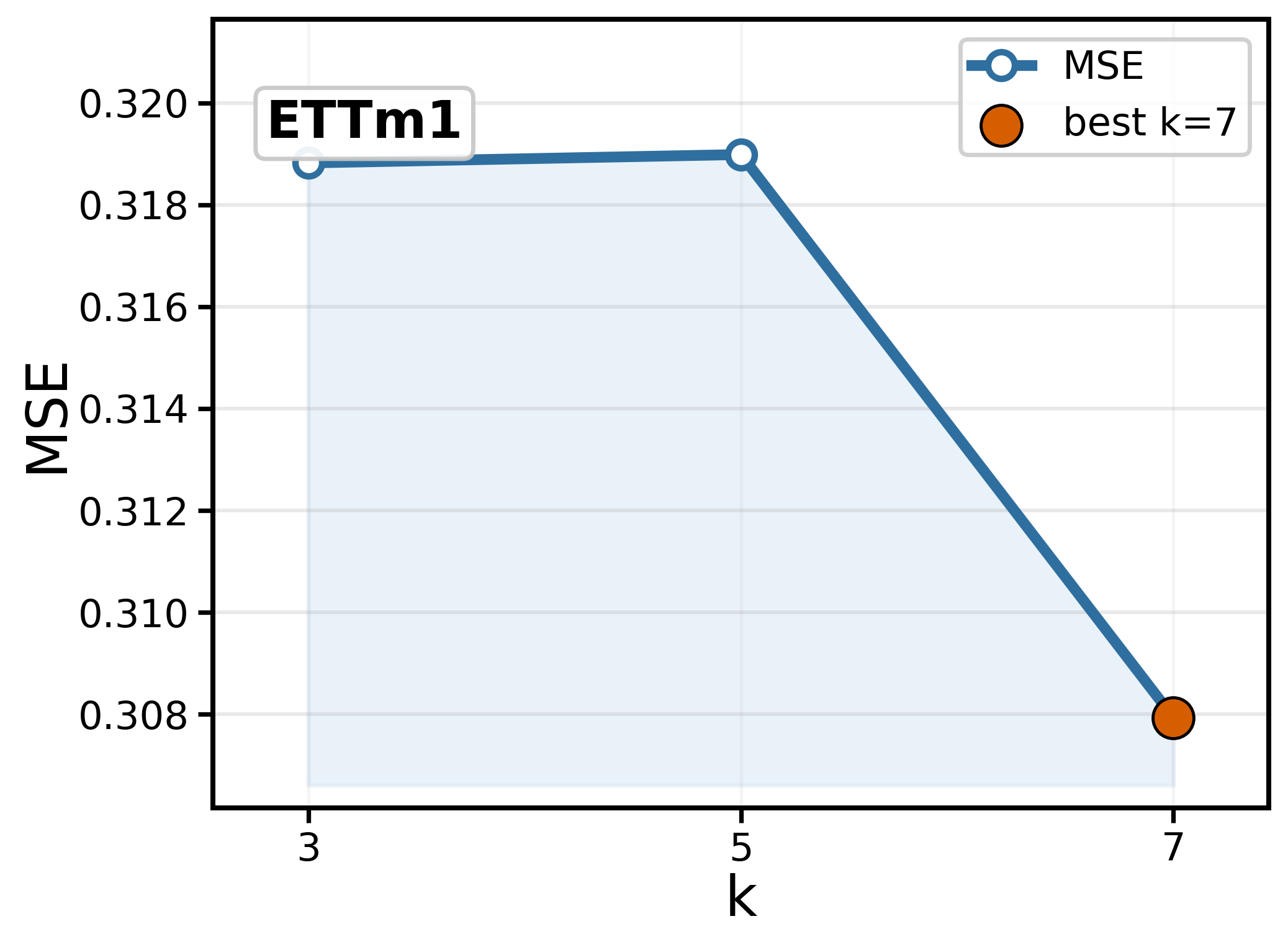}}
\subfigure[\scriptsize ETTm2]{\includegraphics[width=0.24\linewidth]{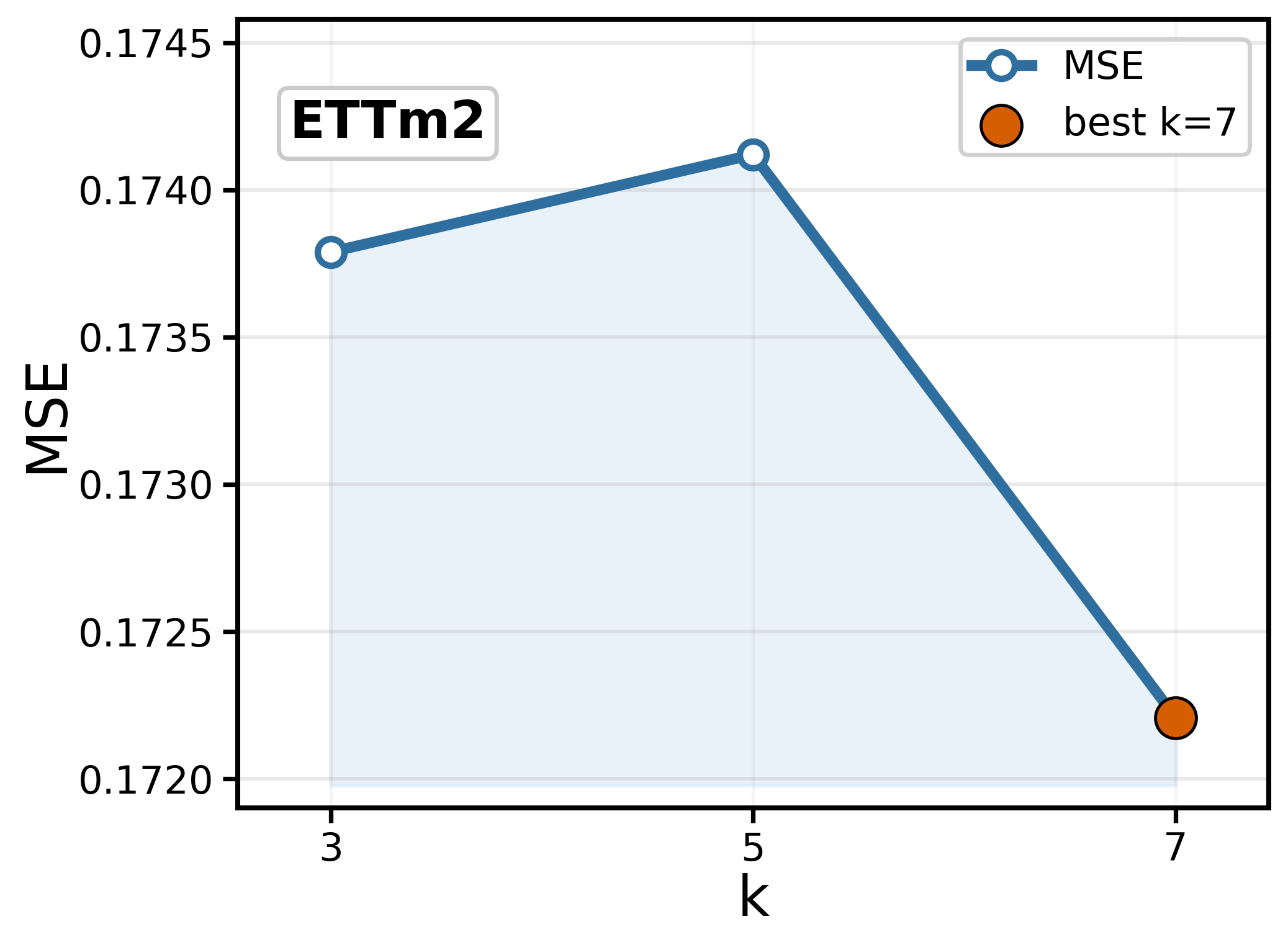}}

\vspace{-1em}

\subfigure[\scriptsize ECL]{\includegraphics[width=0.19\linewidth]{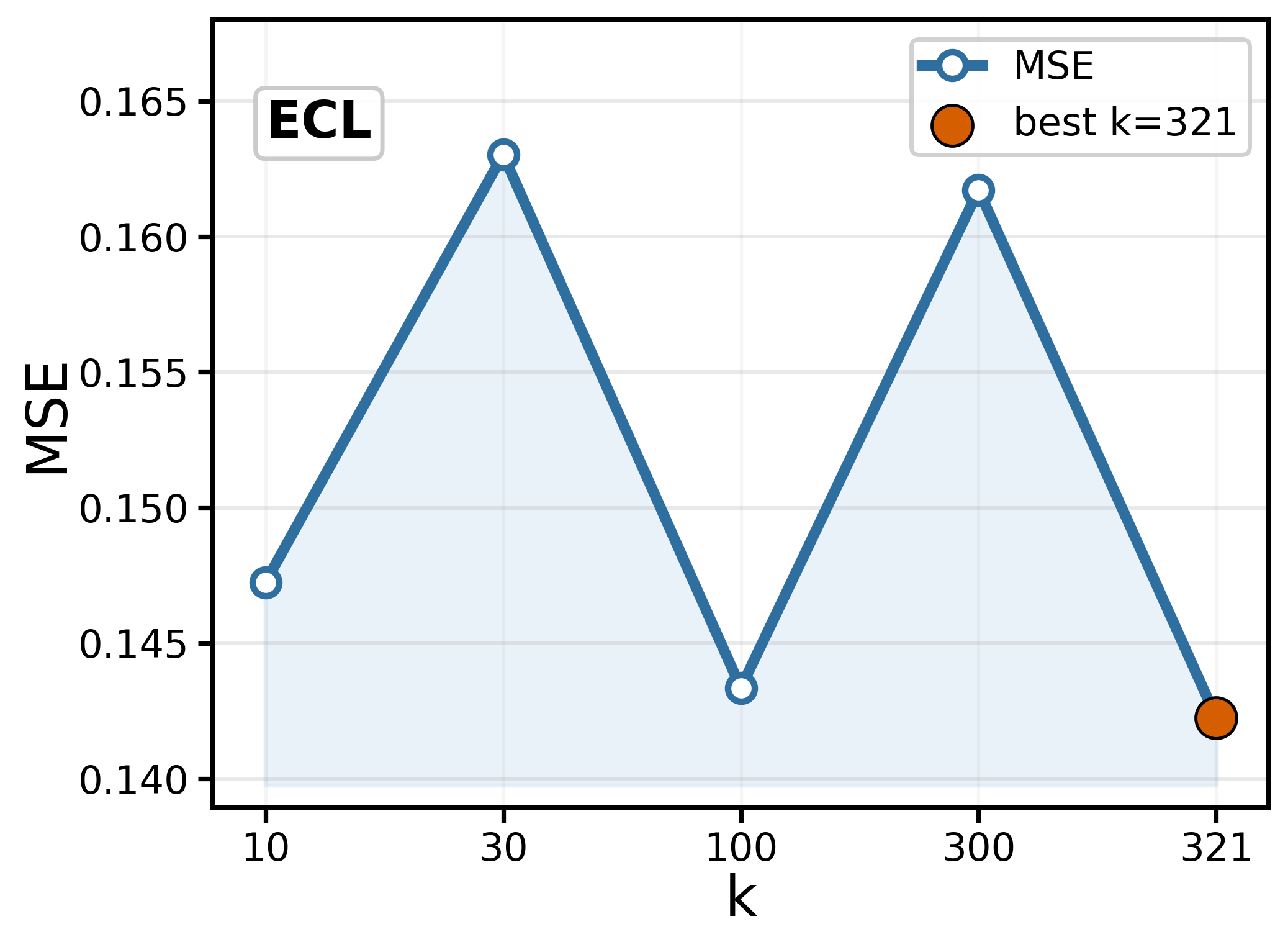}}
\subfigure[\scriptsize PEMS03]{\includegraphics[width=0.19\linewidth]{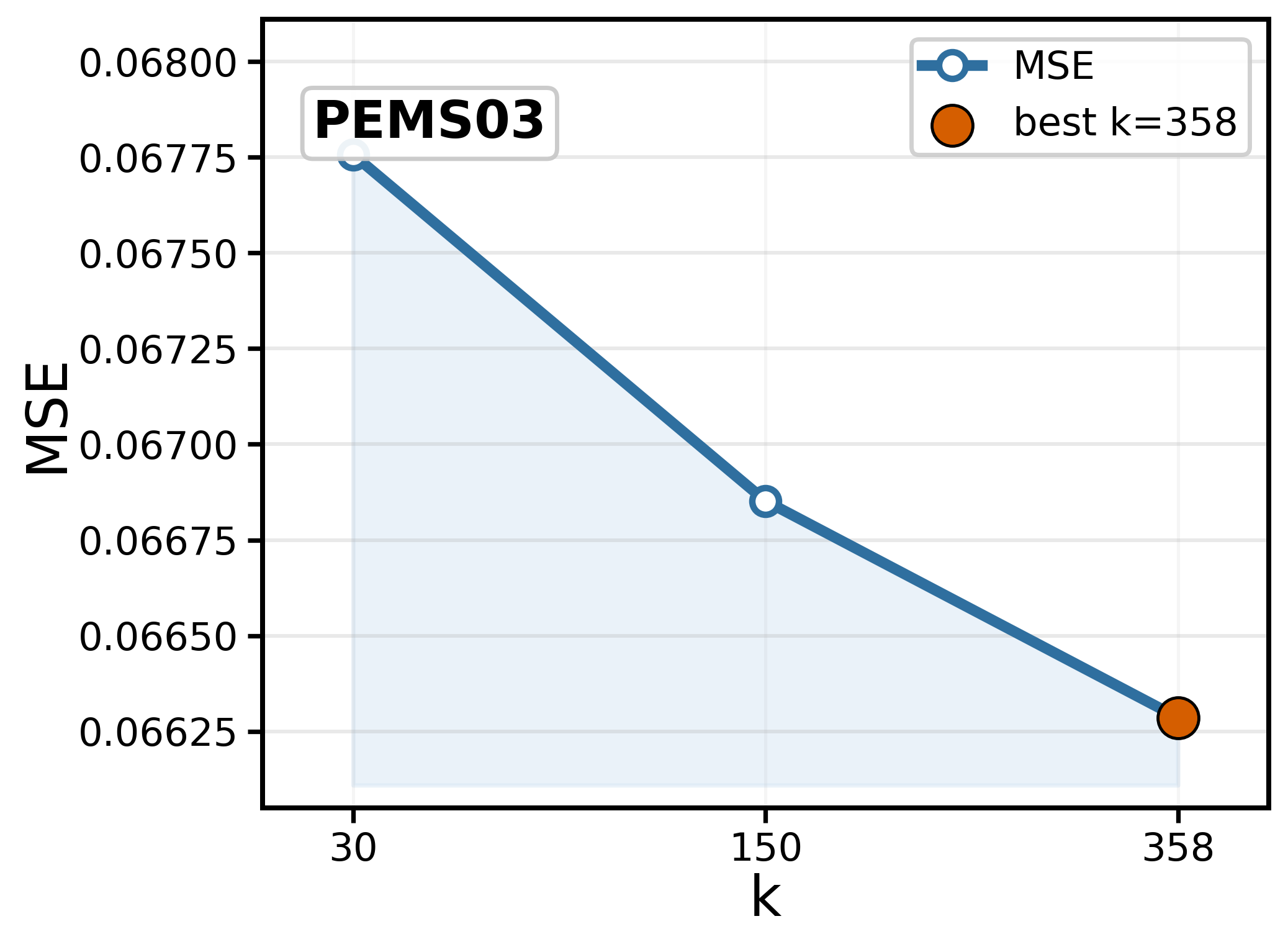}}
\subfigure[\scriptsize PEMS08]{\includegraphics[width=0.19\linewidth]{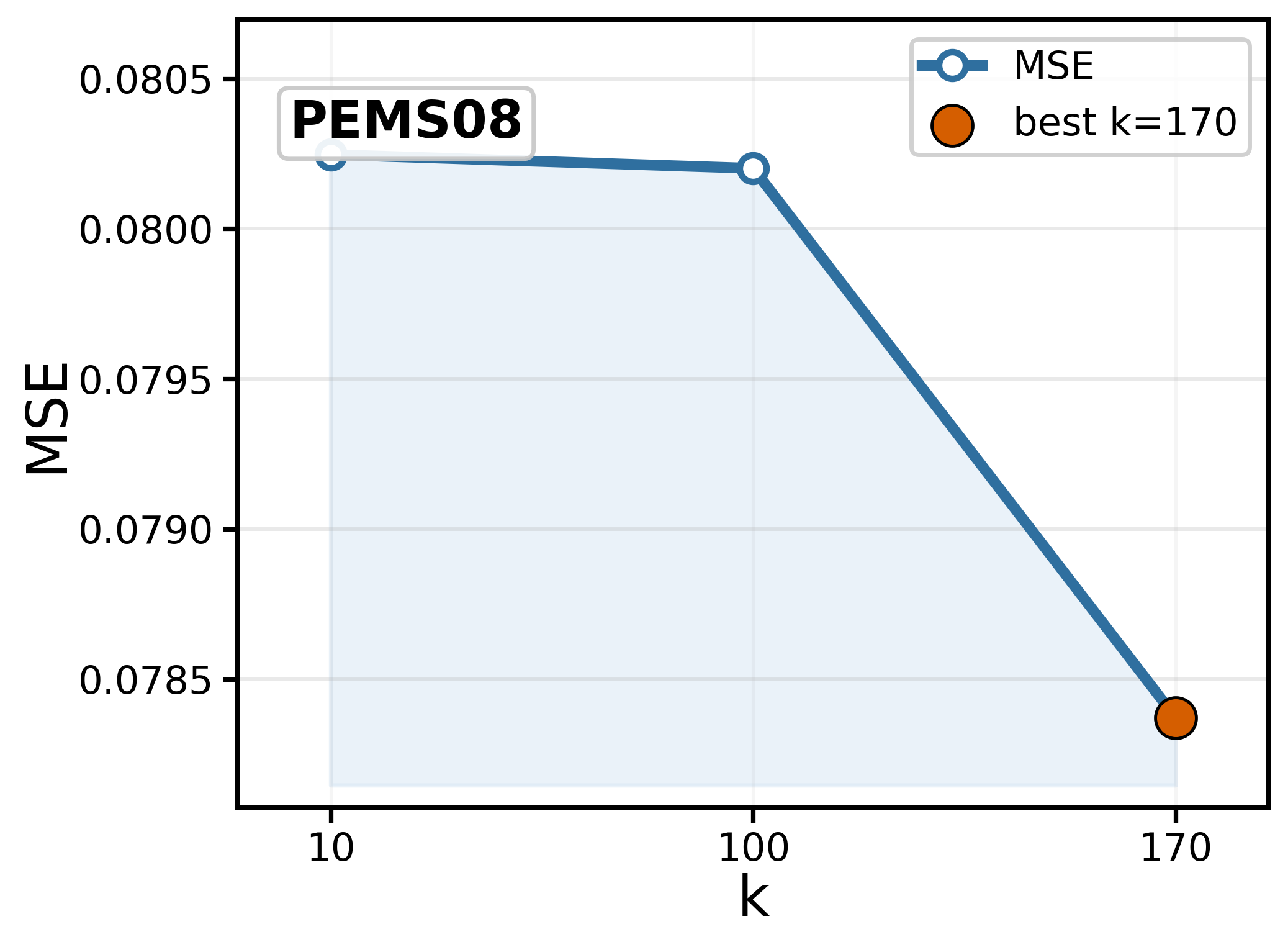}}
\subfigure[\scriptsize Weather]{\includegraphics[width=0.19\linewidth]{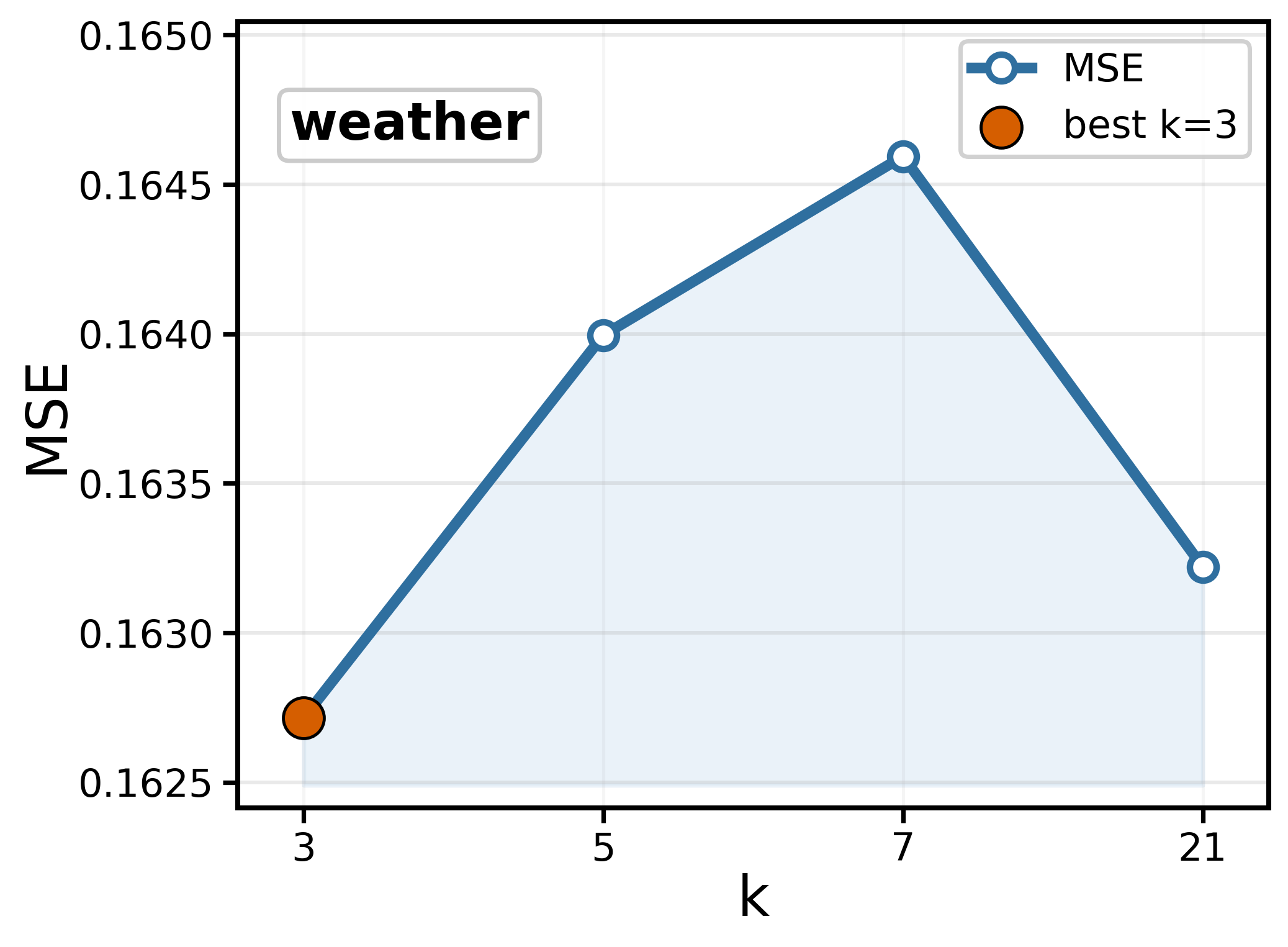}}
\subfigure[\scriptsize Traffic]{\includegraphics[width=0.19\linewidth]{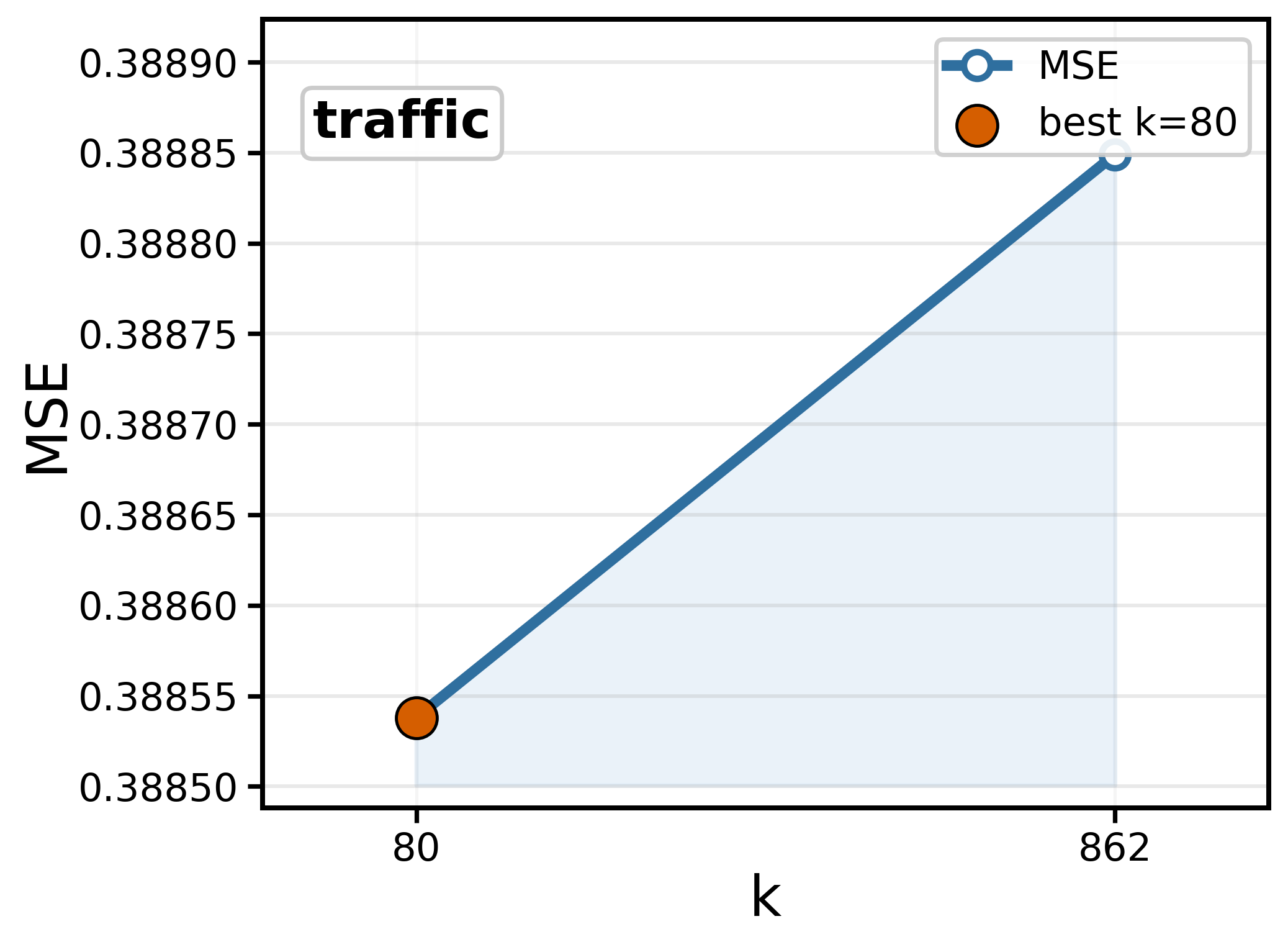}}

\vspace{-0.8em}
\caption{Effect of the number of principal components $k$ across datasets. The orange marker indicates the best $k$ selected by MSE.}
\label{fig:mse_k}
\end{figure}

\paragraph{Sensitivity of PCA dimension $k$.}\label{sec:sensi_k}
We further evaluate the PCA dimension $k$ in Fig.~\ref{fig:mse_k}. CoRe remains stable across a range of values, suggesting that the low-rank projection is not highly sensitive to this choice. Low-dimensional ETT datasets typically prefer most available components, which is expected because only a few variables are present and discarding components can remove useful signal. High-dimensional datasets exhibit more dataset-specific optima: ECL, PEMS03, and PEMS08 benefit from relatively larger subspaces, while Weather and Traffic can perform well with more compact projections. This behavior suggests that the PCA projection preserves dominant relational structure while filtering redundant or noisy cross-variable variation.

\vspace{-1.5em}

\vspace{-1em}
\section{Conclusion}

We introduced \method, a model-agnostic objective for direct multivariate time-series forecasting. Rather than relying solely on pointwise errors, \method\ imposes two complementary structural constraints on the predicted future: frequency coherence alignment for temporal structure and low-rank relational graph alignment for cross-variable consistency. It can be applied to existing forecasting backbones without architectural changes or additional trainable parameters. Experiments on standard benchmarks show consistent improvements across strong baselines and different backbones. Ablation and sensitivity analyses further confirm the complementarity and robustness of the two components, supporting output-space structural supervision as an effective approach to multivariate forecasting. Extending the relational term beyond a linear low-rank basis is a natural next step.

\scriptsize
\bibliographystyle{splncs04}
\bibliography{ref}
\end{document}